\documentclass[journal,twoside]{IEEEtran}
\usepackage{array}
\newcommand{\PreserveBackslash}[1]{\let\temp=\\#1\let\\=\temp}
\newcolumntype{C}[1]{>{\PreserveBackslash\centering}p{#1}}
\newcolumntype{R}[1]{>{\PreserveBackslash\raggedleft}p{#1}}
\newcolumntype{L}[1]{>{\PreserveBackslash\raggedright}p{#1}}

\usepackage{multirow}
\usepackage{hyperref}
\usepackage{graphicx}
\usepackage[table,xcdraw]{xcolor}
\usepackage{setspace}

\usepackage{cite}

\usepackage[ruled,vlined]{algorithm2e}
\usepackage{algorithmic}

\ifCLASSINFOpdf
\else
\usepackage{graphicx}
\fi
\usepackage[cmex10]{amsmath}
\usepackage{amssymb}
\usepackage[caption=false,font=footnotesize]{subfig}
\usepackage{color}    

\begin{document}
%
\title{State-aware Anti-drift Robust Correlation Tracking}
%
%
%

\author
        {Yuqi~Han,~\IEEEmembership{Student Member,~IEEE,}
        Chenwei~Deng,~\IEEEmembership{Senior Member,~IEEE,}
        \\ Zengshuo~Zhang,
        Jinghong~Nan,
        and~Baojun~Zhao
}

%
%



\markboth{Paper under Review}{\MakeLowercase{\textsc{H}AN {\emph{et al.}}}:
State-aware Anti-drift Robust Correlation Tracking}

\maketitle

\begin{abstract}
\boldmath
Correlation filter (CF) based trackers have aroused increasing attentions in visual tracking field due to the superior performance on several datasets while maintaining high running speed. For each frame, an ideal filter is trained in order to discriminate the target from its surrounding background. Considering that the target always undergoes external and internal attributes during tracking procedure, the trained filter should take consideration of not only the external distractions but also the target appearance variation synchronously. To this end, we present a State-aware Anti-drift CF tracker (SAT) in this paper, which joint model the discrimination and reliability information in filter learning. Specifically, global context patches are incorporated into filter training stage to better distinguish the target from backgrounds. Meanwhile, a color-based reliable mask is learned to encourage the filter to focus on more reliable regions suitable for tracking. We show that the proposed optimization problem could be solved using Alternative Direction Method of Multipliers and fully carried out in frequency domain to speed up. Extensive experiments are conducted on OTB-100 datasets to compare the SAT tracker (both hand-crafted feature and CNN feature) with other relevant state-of-the-art methods. Both quantitative and qualitative evaluations further demonstrate the effectiveness and robustness of the proposed work. 

\end{abstract}


\IEEEpeerreviewmaketitle

\section{Introduction}
\label{sec:1}


\IEEEPARstart{V}{isual} tracking is one of the fundamental task in computer vision with a plethora of applications such as video surveillance, robotics, human-computer interaction etc. The goal of visual tracking is to estimate the state of the target (location and size) in subsequent video frames with only the initial target position given. Most of the existing trackers \cite{d42,d43,d44} tackle this issue by exploiting machine learning techniques to train a robust classifier or filter based upon the extracted feature of the target and its surrounding background. With the use of powerful classifiers, such tracking methods have achieved competitive results on both accuracy and robustness. However, considering the time-critical property for tracking application, the efficiency for the aforementioned trackers are limited by the number of training samples.


\begin{figure}[!t]
\vspace{0.07cm}
\centering{
\includegraphics[width=9cm,height=7.cm]{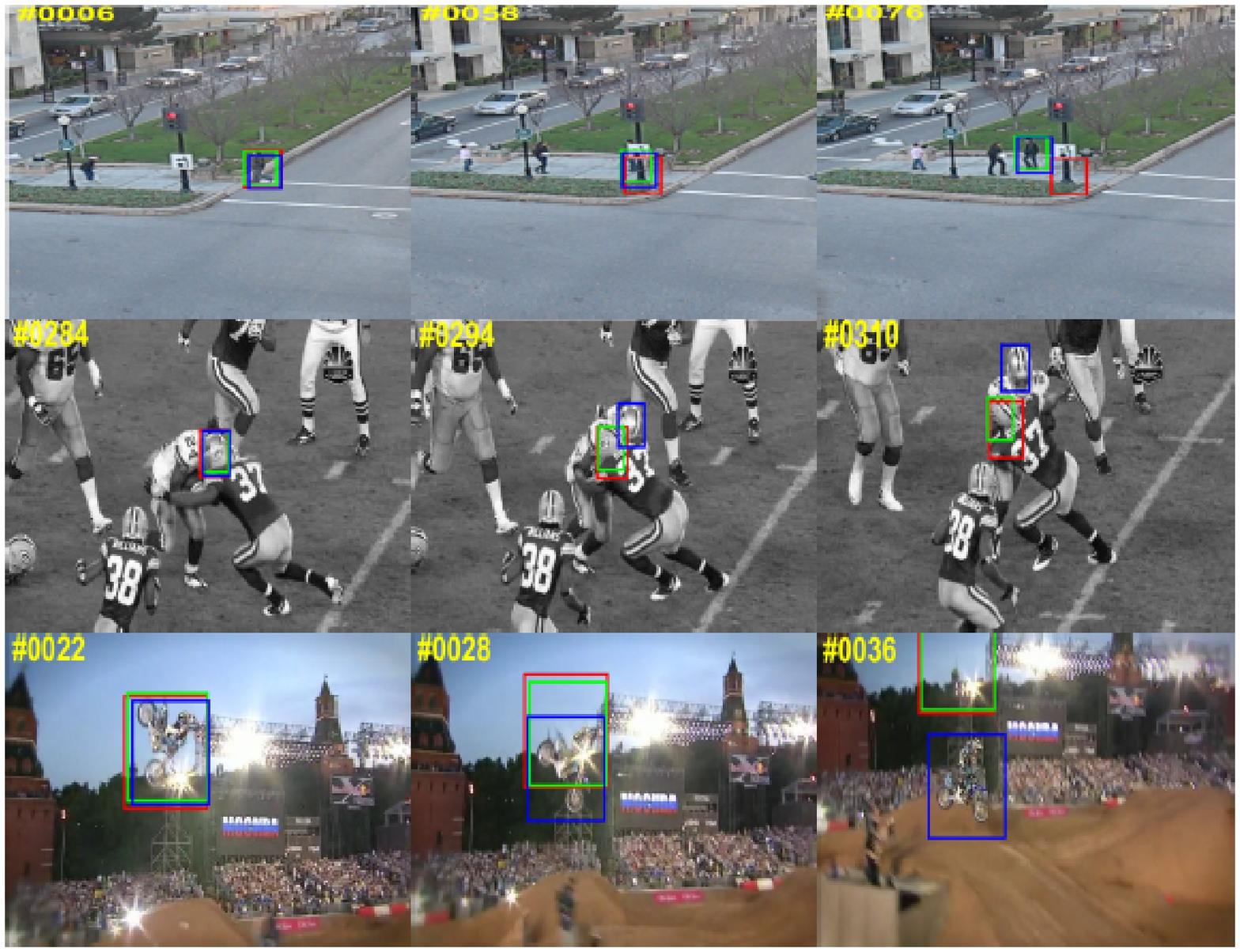}\hspace{-0.2cm} 
}
\\ \vspace{-0.3cm}
\includegraphics[width=5.5cm,height=0.7cm]{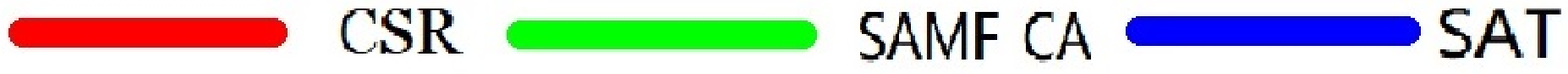}\\ \vspace{-0.4cm}
\caption{ Comparisons of our SAT trackers with state-of-the-art trackers CSR and SAMF-CA in challenging situations of background clutter, deformation and occlusion on sequence Human3, Football and Motorrolling, respectively. Best viewed in color.}
\label{fig:1}
\end{figure}

Recently, Correlation filter (CF) based trackers \cite{d3} have sparked a lot of interest due to their high accuracy while running at high speed. Instead of randomly extract positive and negative samples in a small searching window, CF trackers approximate dense sampling strategy by circularly shifting the training samples. Therefore, instead of solving the computational-cost matrix inversion, CF trackers could handle the issues by element-wise operation in Fourier domain, by taking advantage of the property for circulant matrix. 

Despite of the simplicity and the huge success they have achieved these years, CF trackers still suffer from model drift problem due to the challenging factors in tracking scenarios. Figure 1 illustrates the tracking results of two state-of-the-art CF trackers (CSR\cite{d16} and SAMF-CA \cite{d24}), on several challenging sequences in \cite{d2}. One can see that the CSR tracker fails to locate the target stably when occlusion or similar distractor appears. On the other hand, the SAMF-CA tracker shows inferior performance when the target undergoes rotation as shown in the third row.

In terms of such phenomenon, we notice that even though various of attributes could lead to model drift and tracking failure problem, as listed in OTB13 \cite{d1}. They could be categorized into the external interference and internal target appearance variation. For the external  interference, even the target appearance remains steady while there are distractors which have similar color or texture with the target in the surrounding background. In extreme cases (i.e. occlusion or background clutter occurs), the target appearance may be contaminated by these distractors. On the other hand, in scenarios with challenging deformation, rotation and scale variation, the background texture remains stable while only the target appearance model changes rapidly. Thus, we argue that a robust tracker should be able to handle both the external interference and the internal issues simultaneously. However, most of the existing CF trackers usually fail to pay attention on both sides.

To tackle such limitations, we advocate a novel CF-based optimization problem and further develop a state-aware anti-drift tracker (SAT) in this paper, which jointly model the discriminative and reliable information in filter learning stage. Specifically, surrounding contextual patches are employed into tracking framework in order to equip the tracker with discriminative ability against the external distractions. Furthermore, a color-based reliable mask is learned for each frame to segment the foreground as well as encourage the filter to focus on the reliable region when internal interference occurs. The optimal filter is obtained as the element product of the discriminative filter and the reliable mask. We show that the proposed optimization problem could be solved element-wisely using Alternating Direction Methods of Multipliers (ADMM) in Fourier domain, which is computational efficiency. What's more, by introducing kurtosis to serve as the tracking monitor indicator, a simple yet efficient template updating strategy is incorporated to 
avoid the template contamination as well as maintain the target appearance. We test the proposed tracker on OTB-2015 benchmark \cite{d2} which contains 100 sequences with both the external and internal challenging factors to valid our approach. Experimental results show the proposed SAT tracker perform superiorly against the state-of-the-art methods. 

Our contributions could be summarized as follows:

$\bigstar$ In this paper, we attempt to jointly model the discrimination and reliability  information synchronously into the CF tracking framework. The proposed optimization problem could be solved using ADMM technique in Fourier domain with limited computational burden. 

$\bigstar$ We explore the statistical property of the response map in CF trackers, and a novel high-confidence updating scheme is advocated to avoid template corruption as well as ensure the robust tracking. 

$\bigstar$ Extensive experiments are carried out on tracking benchmarks and demonstrate that the proposed tracking algorithm performs favorably against existing state-of-the-art methods.

The remainder of this paper is organized as follows. Section II presents a short description of the work related to ours. In Section III, the proposed approach is elaborated in details. Section IV describes experimental results and related analysis, while we draw our conclusions in Section V.

\section{Related Works}

Most existing trackers adopt either the generative\cite{d4,d5} or the discriminative \cite{d6,d7} approaches. For generative trackers, an elaborated appearance model is often designed to describe a set of target observations in order to search the best-matching patches for the target. While the discriminative trackers formulate visual tracking as a binary classification problem, which search for the target location that is most distinctive from the background. However, the tracking efficiency  are limited by the number of the training samples.

To address the above issues, significant attentions have been paid to discriminative correlation filter (DCF) based trackers \cite{d3}, which instead minimize the least-square function for all the circular shifts of the positive samples. Although this is the approximation for the actual problem, it enables dense-sampling strategy and further transfers the computational-cost matrix inversion to element-wise operations in Fourier domain. Using correlation filters for tracking started with MOSSE \cite{d8}. Using the single-channel gray features, MOSSE achieved the state-of-the-art performance on tracking benchmark, running at a high speed of more than 600 FPS. Henriques et al. incorporate both non-linear kernels and multi-dimensional features to replace the original grayscale template in \cite{d9}, achieving the state-of-the-art performance in VOT 2014. Seminal follow-up works of DCF have been proposed either in the performance advancement or the conceptual improvements in filter learning. DSST \cite{d10} and SAMF \cite{d11} add an extra scale filter to adapt the scale variations. Muster \cite{d12}, ROT \cite{d13} and LCT \cite{d14} carefully design re-detection schemes for long-term tracking. Color information has been taken into account in CN \cite{d15}, Staple \cite{d16} and CSR \cite{d17} to achieve the tracking robustness to non-rigid deformation. Furthermore, inspired by the recent success of the convolution neural networks (CNN) in object classification, researchers in tracking community devote theirselves on deep trackers, which could take advantage of the robust feature representation of CNN networks. \cite{d18} and \cite{d19} employ pre-trained CNN feature instead of hand-crafted features, and the final results are obtained by stacking hierarchical responses and hedging trackers, respectively. Martin et al. investigate the feature map combination for continuous convolution filter with different spatial resolutions in  \cite{d20}. With fewer super-parameters to tune, the C-COT tracker is insusceptible to the over-fitting problems.  

Except for the forementioned literature which focus on combining the DCF framework with feature representation or detection module. Another research hotspot is dedicated to tackling the inherent limitations of DCF tracking by modifying the conventional CF loss function for filter training. In SRDCF \cite{d21}, a spatial regularization term is added to the basic loss function to alleviate the boundary effect. The BACF tracker \cite{d22} trains the filter with a binary mask, which could generate more real samples as well as maintaining computational efficiency. Similar to BACF, Bibi et al. modify the expected target response and decrease model drift problem significantly in \cite{d23}. Recently, a Context-aware tracker \cite{d24} has been proposed to explicit the global context within CF trackers. Compared to the conventional CF trackers, the CA tracker is adept at suppressing the potential distractions at background regions. However, targets undergo several various of challenging attributes (both the external and the internal ones), during tracking process. Their learned trackers would always tend to be interfered by the salient parts in the feature map due to target appearance change itself. Hence, different from the existing methods, we aim at training the filter with discriminative context and reliable target information jointly, which could adapt to diverse tracking challenges. Furthermore, we explore a novel tracking confidence monitoring criterion to measure both the accuracy and robustness of the predicted results. With the employment of the proposed criteria, our algorithm is able to forecast the potential distractions or temporal tracking failures in time, achieving the state-aware performance. Accordingly, the proposed SAT tracker would adapt the updating strategy, maintaining the stable and purity of the training samples. We test the proposed framework in a general benchmark to validate its effectiveness.

\section{Proposed Tracking Algorithm}
\label{sec:3}

We base our tracking algorithm on three fundamental requirements for online tracking: Firstly, to reduce the risk of drifting in cases of external challenging interference, the tracker should be aware of the potential distractors and have the ability to identify them beforehand. Secondly, the object model should reliably represent the target as well as suppress the background sub-region in the bounding box, when the target undergoes internal appearance variation. The last but not least, the tracker should have the ability to measure the tracking condition, which could further adjust the updating strategy in a high-confidence manner. It's conceivable that such criterion could assist the tracker to recover from model drift  as well as maintaining the purity of learned filter. Therefore, we propose a CF-based optimization function which jointly addressing these requirements by training the filter with discriminative and reliable information. Pleasantly, we show that the optimization problem could be solved using ADMM technique and fully carried out in Fourier domain to speed up. Since the proposed SAT tracker apply the CF tracker as the baseline method, we would revisit the details about conventional CF tracker firstly. Furthermore, we also implement SAT tracker with the deep feature and scale adaption module to validate the strong compatibility of our algorithm.

\subsection{Correlation Filtering}
\label{ssec:3.1}

The goal of discriminative tracking method is to train a classifier or filter which could be applied to the region of interest in consecutive frames to distinguish the target from backgrounds via learning features from positive and negative samples. The optimal classifier is learned as follows:

\vspace{-0.25cm}
\begin{equation}
{\bf w}_{opt} = \arg\min\limits_{\bf w}\sum^N_{i=1}(\sum^D_{d=1}{\bf w}_{d}*{\bf x}_{i,d}-y_i)^2_2+\lambda\|{\bf w}\|_{2}^{2}  
\end{equation}

 $N$ denotes the number of training patches, and $d$ stands for the index of the feature channel.  ${\bf x}$ denotes the input feature and $\bf y$ is the corresponding regression label ranging from one to zero. The forementioned objective function has a global minimum due to its convexity. We could acquire the close-form of the optimal classifier ${\bf w}_{opt} = ({\bf X}^H{\bf X}+\lambda {\bf I})^{-1}{\bf X}^H{\bf y}$, when we gather the feature of all the training samples to form a data matrix ${\bf X}$. For simplification, all derivation would be inferred under single-channel feature condition.  

Due to the computational burden in solving the matrix inversion, most of the previous work randomly picked a limited number samples from the searching region around the target. Such stochastic sampling strategy would bring uncertainty to the tracking performance. To tackle such issues, CF based trackers allow dense sampling scheme within the searching area at low computational cost. The key innovation of this technique is to approximate the spatial exhaustive searching by efficient dot product in the frequency domain taking advantage of the following property \cite{d26}.

Property: We denote the conjugate of the feature vector ${\bf x}$ by ${\bf x}^*$ and the Fourier transform by $\bf{\hat{x}}$. Hence, we could establish the connection between the input vector $\bf x$ with the circulant matrix as: ${\bf X} = {\bf F}diag({\bf {\hat x}}){\bf F}^H$ and ${\bf X}^H = {\bf F}diag({\bf {\hat x}^{*}}){\bf F}^H$. Therefore, we could derive the optimal classifier efficiently in Fourier domain: 
\begin{equation}
\bf{\hat{w}} = \frac{\hat{x}\odot\hat{y}}{\hat{x}\odot\hat{x}^\ast+\lambda}
\end{equation}

After the filter is learned  in the current frame, it would be multiplied with the circulant matrix ${\bf Z}$ of the image patch ${\bf z}$ in the following frames. The detection formula is given as below:

\begin{equation}
\vspace{-0.15cm}
 {\bf S}({\bf w},{\bf Z}) = {\bf Z}{\bf w} \Leftrightarrow {\bf \hat{S}} = {\bf \hat{z}} \odot {\bf \hat{w}}
\end{equation}

The product shares the same size with the searching region, and the maximum score ${\bf S}$ indicates the target location in that frame. Afterwards, the filter is updated using the new object location frame by frame with a learning rate  $\eta$ so as to maintain the historical appearance representation of the target.

\begin{equation}
\vspace{-0.3cm}
{\bf \hat{w}}^t= (1-\eta){\bf \hat{w}}^{t-1}+\eta {\bf \hat{w}}^{t}
\end{equation}

\subsection{Robust Filter Learning}
\label{ssec:3.2}

Albeit their simplicity, CF trackers still suffer from several inherent drawbacks due to the challenging factors in tracking task. On one hand, due to the limitation for tracking application, the trackers have very limited information about the tracking condition and surrounding context, which leads to model drift when external interference attributes such as occlusion and background distractions occur. On the other hand, when the target undergoes several internal appearance variations, such as shape deformation or in-plane rotation, the target model could not be precisely approximated by an axis-aligned rectangle bounding box, especially for the non-grid object. Consequently, the filter would learn from the background inescapably which would further contaminate the training samples and lead to model drift. To this end, we argue that a robust tracking framework should consider both the external and internal challenging factors synchronously. Specifically, the tracker should be aware of the surrounding potential distractions in advance as well as distinguish reliable foreground from spurious background in the bounding box. 

Based on the previous description, we advocate a novel CF-based optimization problem to learn from the discriminative and reliable information and then develop a state-aware tracking method (SAT). The proposed optimization problem is composed of a discrimination term and a reliability term, which would be elaborated subsequently. For simplification, the following derivation would be presented for single feature channel, but it could be extended to multi-channels easily without loss of generality. 

\textbf{Discrimination Modeling.} Different from the conventional CF trackers which only train the filter and detect the target in a small local neighborhood. We incorporate the surrounding contextual information into the tracking framework during filter training stage to equip the tracker with the discriminative ability in forecasting the potential distractions. 

Specifically, $k$ contextual patches surrounding the estimated target patch would be extracted. Subsequently, the feature circulant matrix for the target and the context patches, denoting as ${\bf A}_0 \in \mathbb{R}^2$ and ${\bf A}_i \in \mathbb{R}^2$ would be calculated, respectively. In order to protect the filter from contaminating by the potential distractions in the ambient background, we manually assign the regression label to zero for these contextual patches. For now, the objective training problem has been reformulated as:

\vspace{-0.5cm}
\begin{equation}
{\bf w}  = \arg\min \|{\bf A}_0{\bf w}-{\bf y}\|_{2}^{2}  +\lambda_1\|{\bf w}\|_{2}^{2} + \lambda_2\sum_{i=1}^k\|{\bf A}_i{\bf w}\|_{2}^{2} 
\end{equation}

This equation could be further simplified by stacking the circulant feature matrices $A$ to constitute a new feature matrix ${\bf B}$ with the following substitution that ${\bf B} = \left [ {\bf A_0}, \sqrt{\lambda_2}{\bf A_1},...,\sqrt{\lambda_2}{\bf A_k }\right ]^T$. Meanwhile, ${\bf Y} = \left [{\bf y} , {\bf 0} ,..., {\bf 0} \right ]^T$, indicates the concatenate regression labels for the target and context patches. With the previous definition, the objective function has the form as:

\vspace{-0.2cm}
\begin{equation}
{\bf w} = \arg\min \|{\bf B}{\bf w}-{\bf Y}\|_{2}^{2}  +\lambda_1\|{\bf w}\|_{2}^{2} 
\end{equation}

\textbf{Reliability Modeling.} Based upon above designment, the filter is able to resist the external interference in tracking by learning the surrounding context information ahead of time. However, similar to conventional CF trackers, the aforementioned discriminative tracker is confined to learning a rigid template as well. Therefore, it would suffer from the template contamination problem unavoidably when some internal appearance variation happens, i.e. non-rigid deformation or rotations. We argue that such drawback could be tackled by constructing a reliable feature representation which is insensitive to shape variation and further constrain the filter to exclude background pixels out of the bounding box.

To this end, we attempt to construct a binary reliable mask ${\bf r}$, with the element belongs to $\{0,1\}$, indicating the category (foreground or background) for each pixel based upon the color models. Since we couldn't observe the object pixels directly in the search region, we would model the posterior probability for each pixel belonging to the target accordingly.

Here, we denote ${\bf x}\in\mathbb{R}^2$ as the target location in the search region and $o$ as the object present in the scene. Therefore, the posterior probability $p(o|{\bf x})$ could be represented as the product of $p({\bf x}|o)$ and $p(o)$. Since we consider the target presence probability $p(o)$ obeys uniform distribution for simplicity. Thus, the posterior probability is mainly depend on the likelihood function $p({\bf x}|o)$. In addition, we model the posterior probability specified by color histogram ${\bf H}$ due to its insensitivity to shape variation and rotation. 

With the above discussion, we could derive the confidence map for the posterior probability as follow:

\vspace{-0.55cm}
\begin{eqnarray}
p(o|{\bf x})  &=&  p({\bf x}|o) p(o)\nonumber\\
&=& \sum\nolimits_{z\in{(f,b)}} p({\bf x},{\bf H}^z\mid o) p(o) \nonumber \\
&=& \sum\nolimits_{z\in{(f,b)}} p({\bf x}\mid {\bf H}^z,o) p({\bf H}^z,o)
\end{eqnarray}

Here, ${\bf H}^f$ and  ${\bf H}^b$ indicate the color histogram for foreground and  background, which are calculated from the target patch and $k$ surrounding contextual patches with the estimated target location at last frame $P_{t-1}$, respectively. Therefore, the posterior probability could be obtained via histogram projection technique proposed in \cite{d46}. Afterwards, we could yield a basic binary segmentation mask ${\bf r}$ via applying an adaptive threshold as in \cite{d46}. For more details, we refer to \cite{d46}. 

\begin{figure}[!t]
\centering{
\includegraphics[width=9cm,height=7.4cm]{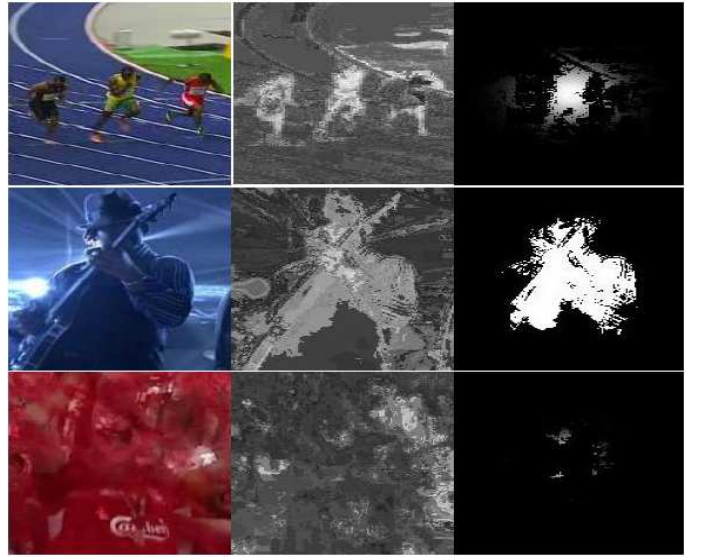}\hspace{-0.2cm} 
}
\\ \vspace{-0.3cm}
\caption{Three examples of the color informativeness test. The first column denotes the current searching region, while the back-projection posterior probability map and binarized segmentation mask are shown in the middle and right column, respectively. The segmentation result in Bolt sequence (the first row) passes the test, while the other two segmentation mask fail the test since too many or too few pixels are assigned to be the target.}
\label{fig:2}
\end{figure}


Since the segmentation mask ${\bf r}$ is obtained based upon standard color histogram, it raises a potential concern that how to avoid the poor classification when the object color is similar to the background or illumination variation occurs. Therefore, we conduct a color informativeness test after obtaining ${\bf R}$. Specifically, we would calculate the deviation between the number of the pixels assigned as foreground with the target size at last frame. Once the 
deviation locates in an ideal range, parameterized by the lower bound $\tau_l$ and upper bound $\tau_u$, we assume that the current color-based segmentation is valid. Accordingly, we would update the color model with a learning rate $\eta_h$. Otherwise, there are too many or too few pixels are labeled to be the target, indicating a potential drastic segmentation failure happens. Under such condition, we abandon the segmentation result for the current frame by setting the mask ${\bf r}$ as all-one matrix and stop the updating for the foreground and background histogram.   

\textbf{Optimization.} Since we have already constructed a reliable mask ${\bf r}$ to tackle the internal interference attributes in tracking procedure. Hence, we could combine the reliable information with the discriminative information jointly in filter learning. To be more specific, we would multiply the trained discriminative filter with the aforementioned reliable mask ${\bf w}_r = {\bf w} \odot {\bf r}$, to encourage the filter to focus on the reliable region and ignore the mixed background area. 

Based upon the above analysis, we construct a novel optimization function which jointly model the discriminative and reliable information of the target and surrounding context patches, resulting in the following Augmented Lagrangian objective function $L({\hat{\bf w}}_{c},{\bf w},{\hat{\bf I}},\rho)$:

\begin{eqnarray}
L({\hat{\bf w}}_{c},{\bf w},{\hat{\bf I}},\rho) = \|{\bf B{\hat {\bf w}_c}-\hat{{\bf Y}}}\|_{2}^{2} + \lambda_{1}\|\hat{\bf w}_{r}\|_2^2  \nonumber\\
+ \hat{\bf I}^T(\overline{\hat{\bf w}_c}-\overline{\hat{\bf w}_r}) + \rho\|(\hat{\bf w}_c-\hat{\bf w}_r) \|_2^2
\end{eqnarray}

Here, we introduce a dual variable ${\bf w}_c$, with the constrain that: ${\bf w}_c-{\bf w}_r \equiv 0$. ${\bf I}$ denotes the complex Lagrangian multiplier and $\rho$ is a positive penalty parameter. Fortunately, the above Augmented Lagrangian function could be solved using the ADMM algorithm \cite{d25} with a series of iterations:

\vspace{-0.3cm}
\begin{equation}  
\left\{  
             \begin{array}{lr}  
             \hat{\bf w}_{c}^{i+1} = \arg\min\limits_{{\bf w}_c} L({\hat{\bf w}}_{c}^{i},{\bf w}^{i},\hat{\bf I}^{i},\rho^{i})  \nonumber \\
             {\bf w}^{i+1} =\arg\min\limits_{{\bf w}} L({\hat{\bf w}}_{c}^{i+1},{\bf w}^{i},{\hat{\bf I}^{i}},\rho^{i}) \\ 
             \hat{\bf I}^{i+1} = \hat{\bf I}^{i} + \rho^{i}(\hat{\bf w}_{c}^{i+1} - \hat{\bf w}_{r}^{i+1} ) \nonumber \\ 
             \rho^{i+1} =  \min(\rho_{max},\beta\rho^{i}) \nonumber 
             \end{array}  
\right.  
\end{equation}  

It should be noted that the convergence of the aforementioned Augmented Lagrangian function could be guaranteed if the penalty parameter $\rho^i$ is non-decreasing and $\sum_{i=1}^{+\infty}\rho^i=+\infty$, according to its theoretical derivation in \cite{d27}. While the stopping criterion for the objective function depends on the residual of the filter in the previous iterations. Once the residual error term $\hat{\bf w}_{c}^{i+1}-\hat{\bf w}_{r}^{i+1}$ is small enough, the optimization process terminated. While after analyzing the experimental results, we find that the residual error drops significantly after the first few iterations. Therefore, we assign the maximum iteration number to five in all the video sequences.

Based upon the discussion above, the closed-form solution for the variable $\hat{\bf w}_{c} $ and $\hat{\bf w}$ could be acquired as:

\begin{eqnarray}
\vspace{0.15cm}
\hat{\bf w}_{c} &=& ({\bf B}^H{\bf B}+\rho)^{-1}(\rho\hat{\bf w}_r + {\bf B}^{H}\hat{{\bf Y}} -\hat{\bf I}^{T}) \nonumber \\
\hat{\bf w} &=& \frac{\sqrt{N}{\bf F}^{H}(\rho\hat{\bf w}_{c}+\hat{\bf I}^T)}{N(\lambda_1+\rho)}
\end{eqnarray}

Recalling that $B$ denotes the stacked circulant feature matrix for the image patch and context patches with the following equation ${\bf B} = \left [ {\bf A_0}, \sqrt{\lambda_2}{\bf A_1},...,\sqrt{\lambda_2}{\bf A_k }\right ]^T$. We could employ the property of the circulant matrix and subsequently re-write the form of the dual variable $\hat{\bf w}_c $ and the jointly learned filter $\hat{\bf w}_r$ into element-wise in Fourier domain as:

\begin{eqnarray}
\hat{\bf w}_c &=& \frac{\hat{a}^{*}_0\odot\hat{y}+\rho\hat{\bf w}_r-\hat{\bf I}^T}{\hat{a}^{*}_0\odot\hat{a}_0+\lambda_2\sum_{i=1}^{k} \hat{a}^{*}_i\odot\hat{a}_i +\rho}  \nonumber \\
{\bf w}_r &=& {\bf r}\odot {\bf w} = {\bf r}\odot \frac{\mathcal{F}^{-1}(\rho\hat{\bf w}_{c}+\hat{\bf I}^T)}{\lambda_1+\rho} 
\end{eqnarray} 

Since all the calculation could be carried out in Fourier domain, the proposed SAT tracker runs at a low computational burden $\mathcal{O}(NlogN)$. For more details about the derivation, please refer to Algorithm 1 and the Appendix  A.

In addition, to counteract scale variation issues, we adopt the same strategy as \cite{d11} to estimate the translation and accurate scale jointly with the trained filter above. Specifically, we set seven searching sizes ranging from 0.94 to 1.06, with the assumption that the target scale wouldn't change significantly between consecutive frames. For more details, we recommend the readers to refer \cite{d11}.

\begin{figure}[!t]
\vspace{-0.5cm}
\centering{
\includegraphics[width=8cm,height=2.1cm]{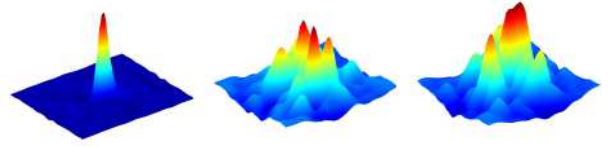}\hspace{-0.2cm} 
}
\vspace{-0.4cm}
\caption{ Illustration of the representative response map during tracking. The response map follows Gaussian distribution under ideal condition as shown in the left column. When the target undergoes appearance change or other external interferences, the response map always shows multiple peaks or abnormal shape as shown in the middle and right column.}
\label{fig:3}
\end{figure}
\vspace{-0.3cm}


\subsection{High-Confidence Filter Updating}
\label{ssec:3.4}
\vspace{-0.1cm}

A recurring question in visual tracking is how to update the target's appearance model so that it could maintain a good representation of the target. Wang et al. \cite{d45} point out that such problem is kind of a stability-plasticity dilemma. Since the tracker must maintain a tradeoff between adapting to new but possibly noisy examples collected and preventing the tracker from drifting to the background. However, implementation of the model updater is often treated as an engineering trick even though their impact on performance is usually quite significant. In this section, we tackle such issues by advocating a novel monitoring criterion to reveal the tracking condition as well as guarantee an accurate and stable filter updating. 
 
Physically, the horizontal axis and vertical axis in the response map indicate the candidate location, while the value ${\bf s}$ could be interpreted as the feature similarity between the target template and candidate samples. Ideally, the response map is assumed to follow a Gaussian distribution, with a single, shark peak and slight tail in the whole searching window, since they are trained with Gaussian shaped regression labels. Unfortunately, under the influence of some challenging attributes and sample noises, the contaminated candidate samples couldn't match the template perfectly, resulting multiple peaks and abnormal shape as shown in Figure 3. Most of the existing CF trackers update their model at each frame without considering whether the detection is accurate. Hence, such trackers always fail to locate the target precisely when the response map is no longer ideal and could hardly recover from the drifting since the filter is contaminated due to the incorrect updating. To this end, we argue that a robust tracker demands not only accurate and stable filter learning but also timely abnormality detection and high-confidence updating strategy as well. 

As mentioned above, the ideal response map should have only one sharp peak and be smooth in other areas. Therefore, the proposed criterion should consider the maximal value in the response map and the distribution of other response value simultaneously. The former attribute could be denoted by the maximum score ${S}_{max}$ in current response map ${\bf S}({\bf Z}, {\bf w})$. For the later measurement criteria, we introduce Kurtosis to measure the peakedness and tail weight of the response distributions. Given a random variable $x$, the kurtosis of the $x$ is denoted as the quotient of fourth cumulant and the square of the second cumulant, which could simply to the fourth central moments minus three. Here the stands for the $k$th cumulant function and $\mu_{k}$ is the $k$th central moment.

\begin{equation}
BK(x) = \frac{\kappa_4(x)}{\kappa_2^2(x)} =\frac{\mu_4(x)}{\sigma^4(x)}-3
\end{equation}

On the basis of the definition, data with high kurtosis tend to have a distinct peak declining rapidly and have heavy tails. While the data with low kurtosis tend to have a flat top or multiple peaks rather than a single sharp peak as illustrated in Fig.3. Based on such mathematic property, we could employ kurtosis to forecast and supervise the tracking quality in advance. 

In addition, we save the kurtosis $BK$ and ${S}_{max}$ and calculate the historical average values each frame. Consequently, we multiply the average values with certain ratio ${\theta_1}$ and ${\theta_2}$ respectively serving as the updating thresholds $S_{tr}$ and $BK_{tr}$.    

\vspace{-0.0cm}
\begin{equation}  
\left\{  
             \begin{array}{lr}  
            S_{tr} = \theta_1 \times \frac{\sum_{t=1}^{T}S_{max}(t)}{T} \nonumber \\
            BK_{tr} = \theta_2 \times \frac{\sum_{t=1}^{T}K(t)}{T}  
             \end{array}  
\right.  
\end{equation}

Therefore, model updating is only performed if both two criterions are larger than the corresponding thresholds with a learning rate $\eta_c$ for the learned filter as in equation 4. Figure 4 illustrates the necessity of the proposed monitoring indicator. The green box indicates the tracking result of trackers which only consider the maximal value while ignore the shape of the  response map as updating metric. The red one stands for the tracking result of our tracker which take both the maximal response and the response shape into consideration during updating. One can see that due to the employment of Kurtosis, inaccurate update is avoided when other distractors appear.

\vspace{-0.4cm}
\begin{figure}
\vspace{-0.2cm}
\centering
\includegraphics[width=9cm,height = 5.2cm]{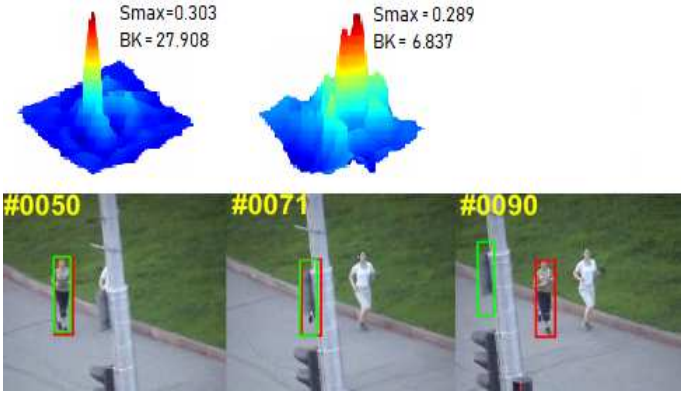} \hspace{-2cm} 
\vspace{-0.35cm}
\caption{Validation results for the proposed high-confidence updating strategy. The first row shows the response maps for frame 50 and frame 71 in Jogging1 sequence. One can see that when the person is occluded by the telegraph, the maximal response still keeps large while the kurtosis value decreases sharply. Hence, with the employment of kurtosis, the unwanted updating would be avoided reasonably. The last picture of the second row illustrates the tracking results when the occlusion ends. The red box could no longer locate the target since the filter is contaminated at frame 71.}
\label{fig:4}
\end{figure}

\begin{algorithm}[t]
\caption{SAT Tracking Algorithm. }%
\LinesNumbered 
\KwIn  {Current Image ${\bf I}_t$, Previous Position ${P}_{t-1}$, Previous Target Size $s_{t-1}$, Learned Filter ${\bf w}_{t-1}$, Previous Color Histogram ${\bf H}_{t-1}$.}
\KwOut{Estimated Target Position $P_t$ and Scale $s_t$, Updated Filter ${\bf w}_t$ and Histogram ${\bf H}_t$ }
\textbf{Repeat:}\\ 
Extract target patch's feature ${\bf a}_0$ and context's features ${\bf a}_i$. \\ 
Extract the foreground and background histogram ${\bf H}^{f}$ and ${\bf H}^{b}$ based on the previous location $P_{t-1}$.\\ 
Construct the reliable mask ${\bf r}$  based on ${\bf H}^{f}$ and ${\bf H}^{b}$. \\
\If{ the color informativeness test passes,}{
 Update the foreground and background histogram:  ${\bf H}^{f}_{t} = (1-\eta_h) {\bf H}^{f}_{t-1}+ \eta_h {\bf H}^{f}$, ${\bf H}^{b}_{t} = (1-\eta_h) {\bf H}^{b}_{t-1}+ \eta_h {\bf H}^{b}$.\\
\textbf{endif} 
}
Train the filter ${\bf w}$ based on the reliable mask ${\bf r}$ and the discriminative features ${\bf a}_0$ and ${\bf a}_i$. \\
Estimate the current target position $P_{t}$ and target size $L_s(t)$ by computing response map. \\
Calculate the maximal response $S_{max}$ and the Kurtosis $BK$ based on the response map. \\
\If{the update condition is satisfied,}{
Update the optimal filter: ${\bf w}_t = (1-\eta_c){\bf w}_{t-1} + \eta_c {\bf w}$. \\
Update the current scale size:  $s_t  = s_{t-1} \times L_s(t)$   . \\
Update the updating threshold $S_{tr}$ and $BK_{tr}$. \\ 
\textbf{end if} 
}
\textbf{Until} the end of video sequence.
\end{algorithm}

\subsection{DeepSAT Tracker}
\label{ssec:3.5}

Recently, with the great power in feature representation, convolutional neural networks have demonstrated state-of-the-art results on a wide range of computer vision task. Therefore, we introduce a pre-trained CNN feature into the proposed framework. Inspired by HCF \cite{d18}, we utilize conv3, conv4, conv5 in VGG-Net as feature extractor. The feature in earlier layer retain higher spatial resolution for precise location, while feature in latter layer capture more semantic information and less fine-gained spatial details. In order to integrate the features in different layers effectively, each layer is convolved with dual correlation to generate a response map. After a resizing process, a final response map is obtained by stacking all the response maps with different weights. It should be mentioned that, in contrast to the work of DLT \cite{d30} and DeepTrack \cite{d31} which update the appearance models by fine-tuning CNNs online, HCF and our DeepSAT tracker use classifier learning for model update, which is computational efficiency.

In addition, we find that the pre-trained CNN features is of limited effectiveness in estimating the target scale but with high computational cost. Hence, we incorporate the scale estimation method proposed in \cite{d10} to tackle the scale issues for DeepSAT tracker. Instead of searching the translation and scale jointly, we would introduce a one-dimensional convolutional filter with HOG feature after locating the target using Deep-CNN features. To be more specific, $W_t$ and $H_t$ denote the width and height of the current target. $L_s$ stands for the layer of the filter and $\alpha_s$ is the scaling parameter. During scale adaption for DeepSAT, a set of patches centered at the current position would be extracted whose size is $\alpha_s^n W_t\times \alpha_s^n H_t$. Here, $n \in \{[\frac{-(L_s-1)}{2}], ... , [\frac{(L_s-1)}{2}]\}$. Afterwards, the response map of each cropped image could be computed as in equation 3. The index $n$ which gives the maximum response is chosen as the accurate scale at the current frame. Please refer \cite{d10} for more details.
Similarly, the scale estimation is only performed when the updating condition is satisfied to speed up. Experiments on \cite{d2} valid such simple strategy in scale estimation.

\vspace{-0.03cm}
\section{Experimental Results and Analysis}
\label{sec:4}
  In this section, we evaluate our tracker on challenging sequences provided on Online Tracking Benchmark \cite{d2}, which involves 11 common challenging attributes. The proposed tracker is compared with 15 representative sota tracking methods in recent years. These trackers could be broadly categorized into two classes: (i) conventional CF based tracker and its variants including KCF \cite{d9}, DSST \cite{d10}, SAMF \cite{d11}, MUster \cite{d12}, LCT \cite{d14}, ROT \cite{d13}, Staple \cite{d17}, SAMF-CA \cite{d24} and CSR \cite{d16}. For instance, DSST and SAMF address the scale variation, while ROT, MUster, LCT aim at tackling occlusion issues. (ii) other representative trackers reported in OTB benchmark or VOT challenges: Struck \cite{d6}, SCM \cite{d4}, TLD \cite{d7}, MEEM \cite{d28} and TGPR \cite{d29}. It should be noted that since feature plays crucial role in visual tracking. For fair comparison, we equip the SAT tracker with hand-crafted features (HOG and CN) when comparing with other trackers.

\begin{table*}
\scriptsize
\centering
\caption{The Overlap Rate and Precision Scores (in percentage) over OTB100 for DeepSAT and the other 8 CNN-based trackers.}
\label{Tab:1}
\renewcommand\arraystretch{1.5}
\begin{tabular}{|c|c|c|c|c|c|c|c|c|c|}
\hline
 & \multicolumn{1}{c|}{\textbf{OUR}} & \multicolumn{1}{c|}{CNN-SVM \cite{d32}} & \multicolumn{1}{c|}{CNT \cite{d33}} & \multicolumn{1}{c|}{HCF \cite{d18}}  & \multicolumn{1}{c|}{CFNet\cite{d34}} & \multicolumn{1}{c|}{Siamaese \cite{d35}}  & \multicolumn{1}{c|}{HDT \cite{d19}}  & \multicolumn{1}{c|}{DeepSRDCF \cite{d36}}  & \multicolumn{1}{c|}{SRDCFdecon \cite{d37}} \\  \cline{1-10}

{Overlap} & \multicolumn{1}{c|}{\textbf{64.3}}  & \multicolumn{1}{c|}{55.4} & \multicolumn{1}{c|}{54.5}  & \multicolumn{1}{c|}{56.2}  & \multicolumn{1}{c|}{56.8} & \multicolumn{1}{c|}{59.2} & \multicolumn{1}{c|}{65.4} & \multicolumn{1}{c|}{63.5} & \multicolumn{1}{c|}{62.7} \\ \hline

{Precision} & \multicolumn{1}{c|}{\textbf{86.4}}  & \multicolumn{1}{c|}{81.4} & \multicolumn{1}{c|}{72.3}  & \multicolumn{1}{c|}{83.7}  & \multicolumn{1}{c|}{74.8} & \multicolumn{1}{c|}{77.3}  & \multicolumn{1}{c|}{84.8}  & \multicolumn{1}{c|}{85.1}  & \multicolumn{1}{c|}{82.5}\\ \hline
\end{tabular}
\end{table*}

\begin{figure*}[!t]
\centering
\includegraphics[width=9cm,height=7.8cm]{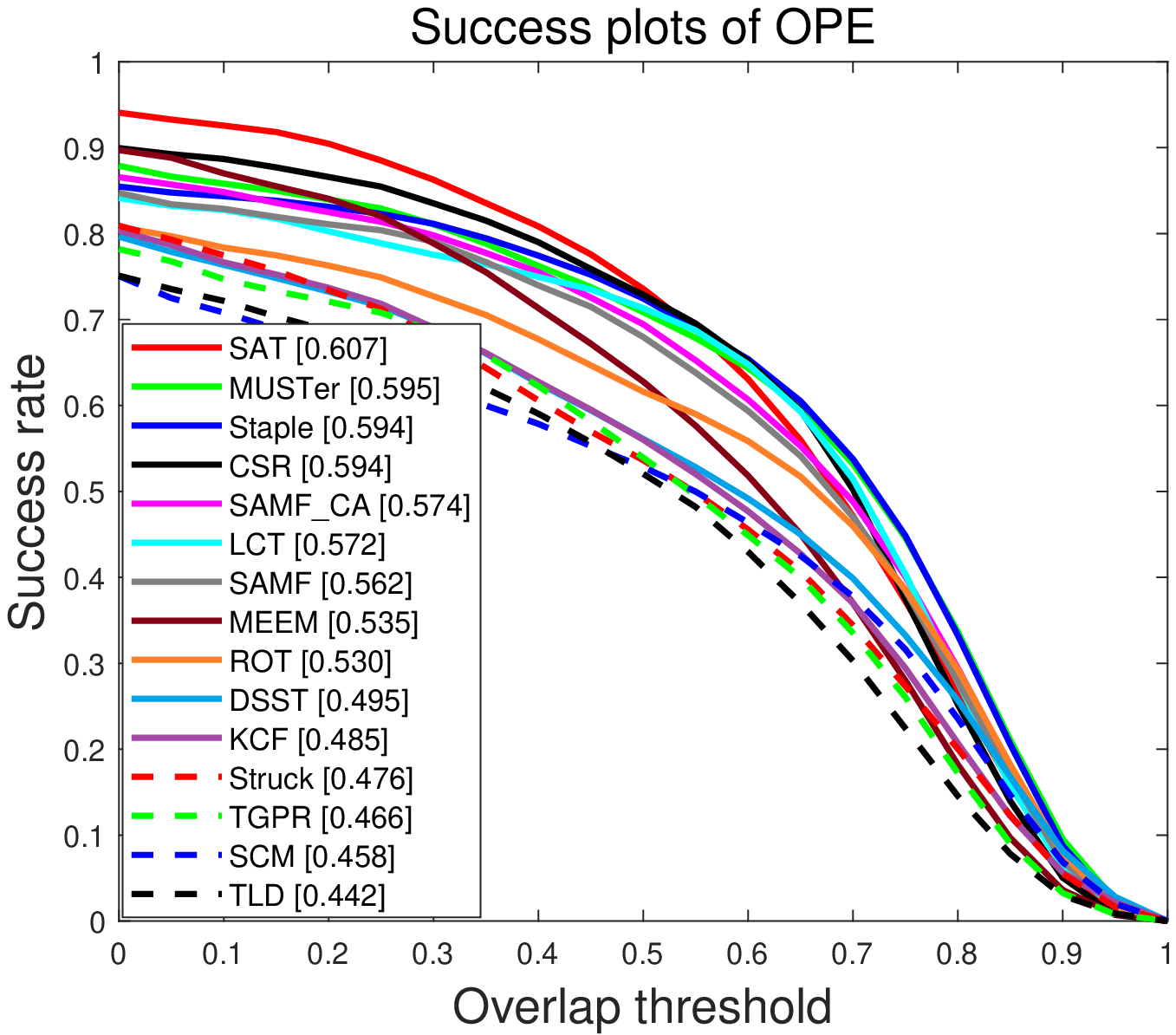}
\hspace{-0.4cm}
\includegraphics[width=9cm,height=7.8cm]{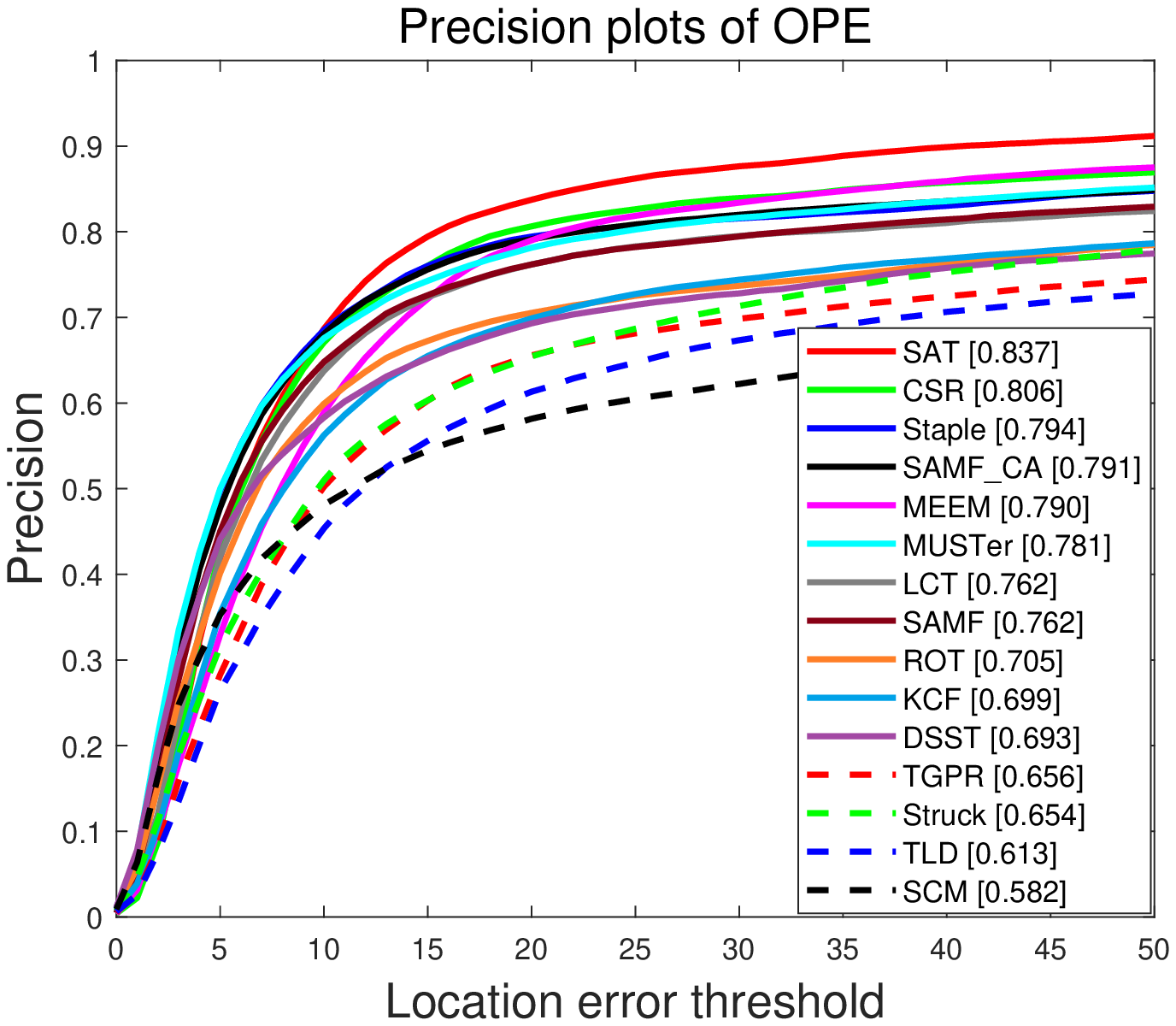}
\hspace{1cm}
\vspace{-0.34cm}
\caption{The success plot and precision plot of OPE. The proposed tracker is compared with 13 state-of-the-art trackers on 100 challenging sequences. The scores of success and precision plots are the values shown in the legend. Best viewed in color.}
\label{fig:8}
\end{figure*}

\subsection{Experimental Setup}
\label{ssec:4.1}
We implement all the experiments in MATLAB 2015a on an Inter(R) Xeon(R) 2.67 GHZ with 32GB RAM. For all the compared trackers, we use the original parameters and source code provided on OTB or the author's websites. HOG and ColorName are selected as the target feature representation for SAT tracker. As for DeepSAT tracker, we exploit an ensemble of deep feature as HCF \cite{d18} (conv5, conv4 and conv3 form VGGdeep-Net). The weight for stacking the response are set to 1, 0.5 and 0.02 respectively. HSV foreground and background color histograms with 16 bins per channel are used to establish the reliable mask for colorful source images. 4 context patches around the target are extracted to boost the discriminative ability of the filter. When conducting the informativeness test, the lower bound $\tau_l$ equals 0.3, while the upper bound $\tau_u$ equals 1.5. The padding size is set to 2.5 times of the initial target size, while the Gaussian kernel width $\sigma$ is set to 1 or 0.25 up to the aspect ratio of the target. The learning rate $\eta_{c}$ for CF template is set to 0.015, while the one for histogram adaption $\eta_{h}$ is set to 0.04. The regularization parameter $\lambda_{1}$ equals to 0.01, while $\lambda_{2}$ is set to 25. During solving the Augmented Lagrangian function, $\rho$ and $\beta$ are fixed to 5 and 3, respectively. The maximum iteration number is set to 5 and the upper bound of the penalty parameter $\rho_{max}$ is set to 25. As for the high-confidence updating section, the threshold ratio $\theta_1$ and $\theta_2$ are set as 0.6 and 0.5, severally. When addressing the scale variation attribute, we adopt two different strategy in SAT tracker and DeepSAT tracker since the pre-trained VGG feature perform limited effectiveness in estimating the target scale. Hence, for SAT tracker, we utilize the scale estimation scheme as in SAMF with 7 search sizes. While we empirically set the number of target pyramid layers $L_s$ to 33, with the scale factor $\alpha_s$ equals 1.02 as in \cite{d10} for the DeepSAT tracker. 

\subsection{Evaluation Methodology}
\label{ssec:4.2}

In this subsection, we employ One-Pass-Evaluation, which is a common evaluation methodology used in Object Tracking Benchmark \cite{d1,d2} to measure the tracking accuracy and robustness of the proposed method against other ones. Two metrics (precision and overlap rate) are utilized to evaluate the performance of candidate trackers. The precision plot illustrates the percentage of the frames whose center locations are within the given threshold distance to the center of the ground-truth. In the experiment, 20 pixels is used to rank the trackers. The success plot is based on the overlap ratio, which is defined as $R = Area (B_T\bigcap B_G)/ Area(B_T\bigcup B_G)$, $B_T$ stands for the tracking output box and $B_G$ stands for the ground-truth rectangle. The success plot shows the percentage of the frames with $R>th$ throughout all the thresholds $th$ belongs to [0,1]. The area under the curve (AUC) of each success plot serves as the second measure to rank the tracking algorithms. Both the precision and success plots show the mean scores over all the sequences.

\begin{figure*}[!t]
\centering
\includegraphics[width=6cm,height=5.2cm]{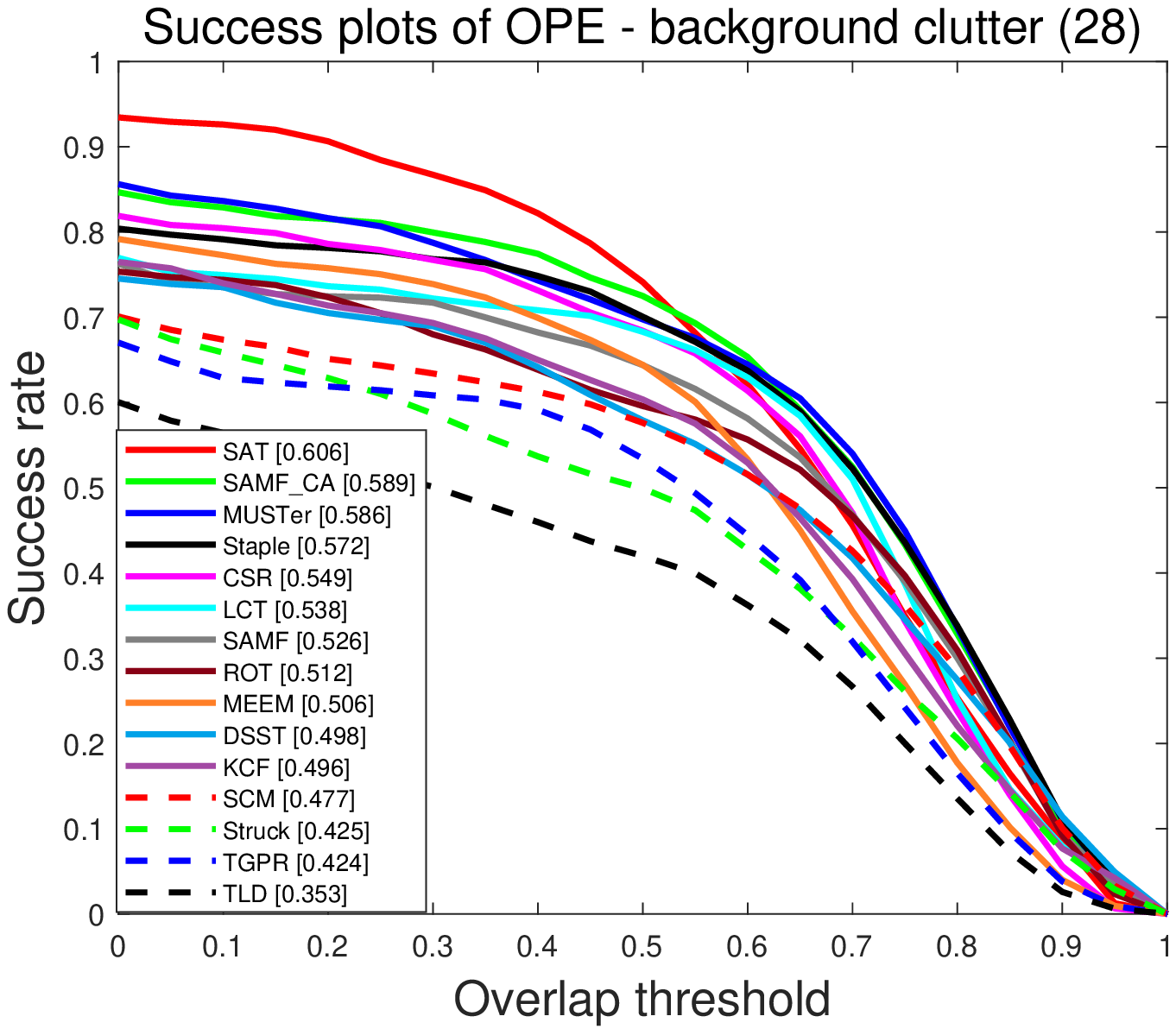}
\hspace{-0.5cm}
\includegraphics[width=6cm,height=5.2cm]{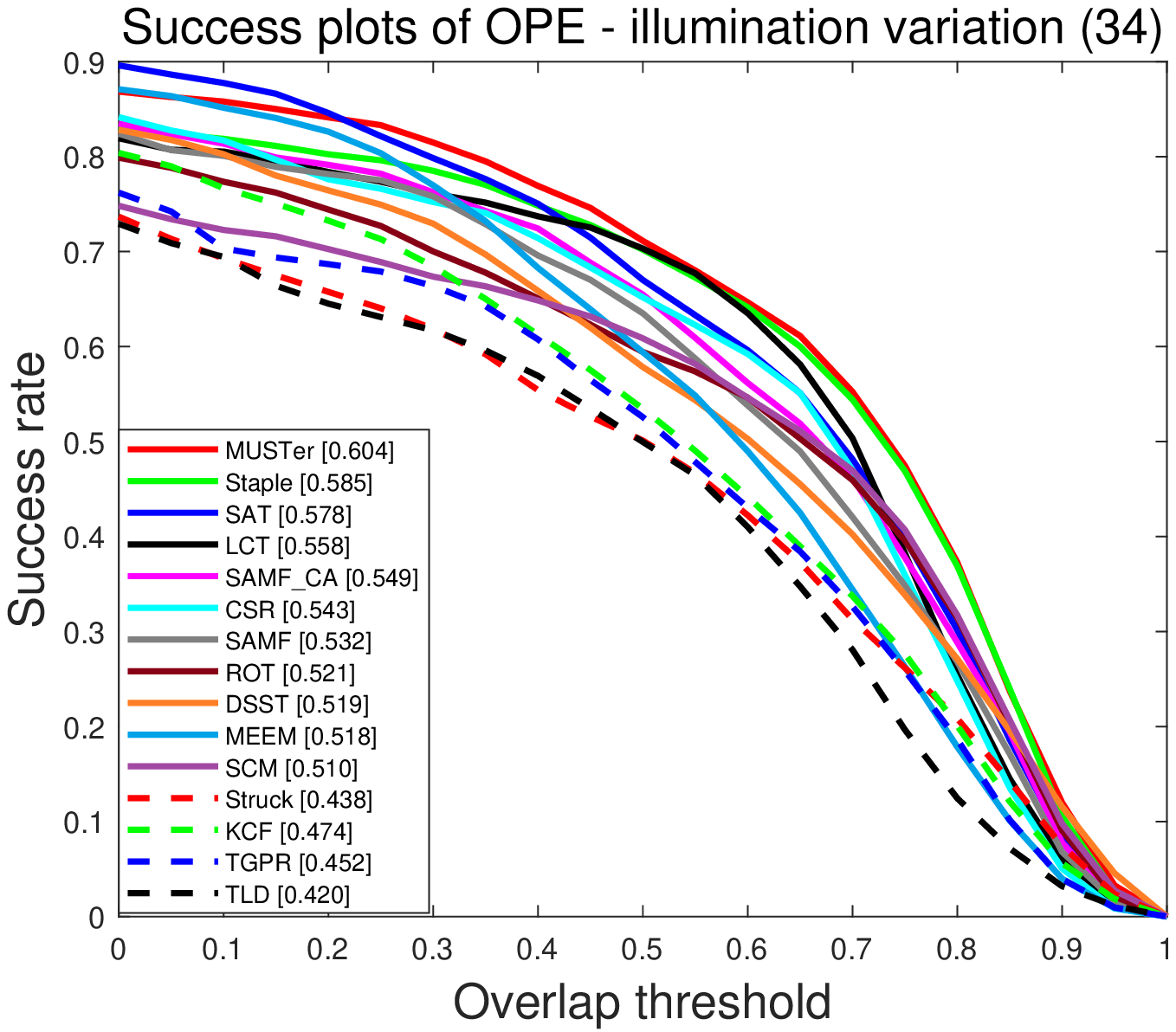}
\hspace{-0.5cm}
\includegraphics[width=6cm,height=5.2cm]{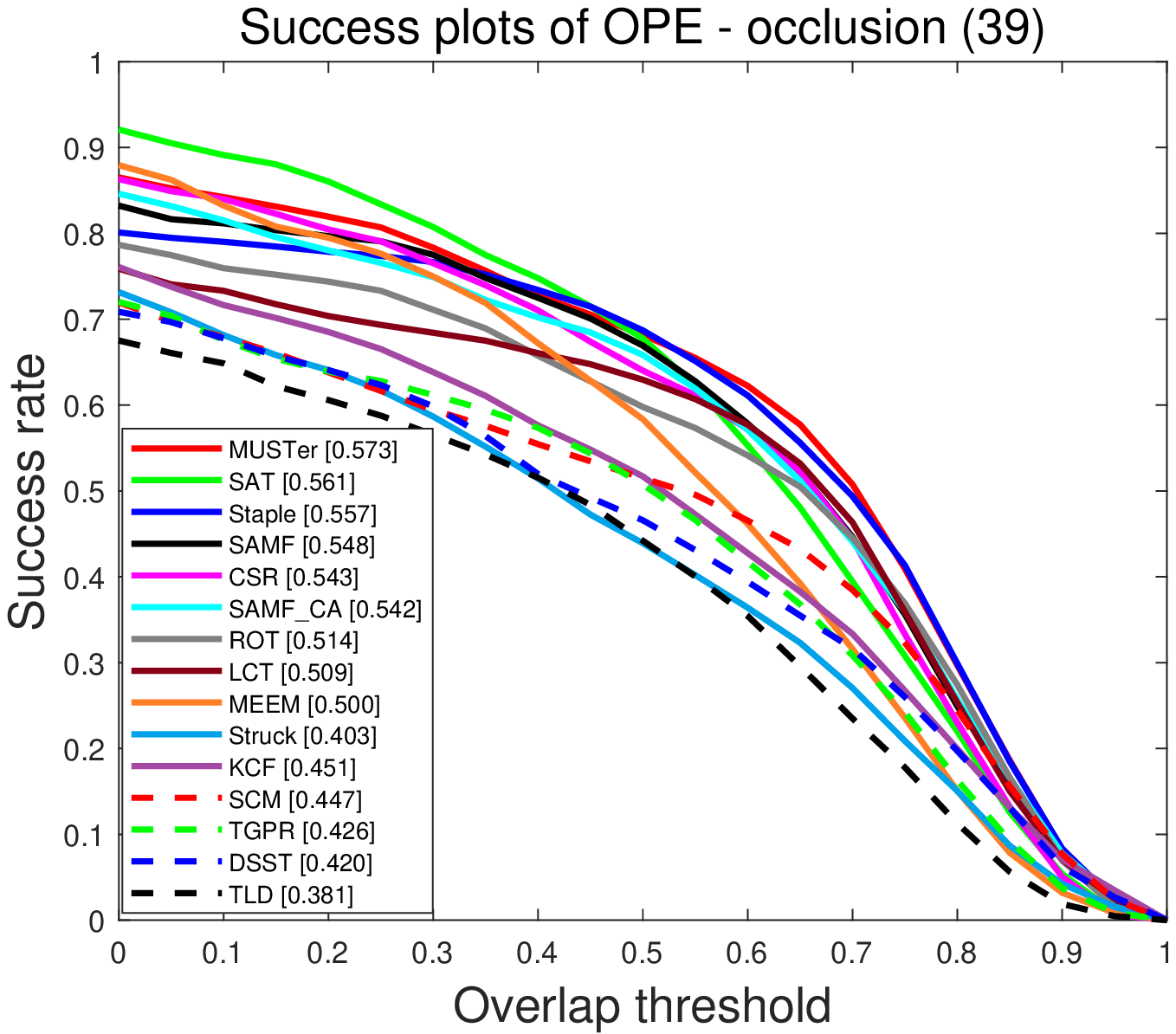} \\ \vspace{0.25cm}
\includegraphics[width=6cm,height=5.2cm]{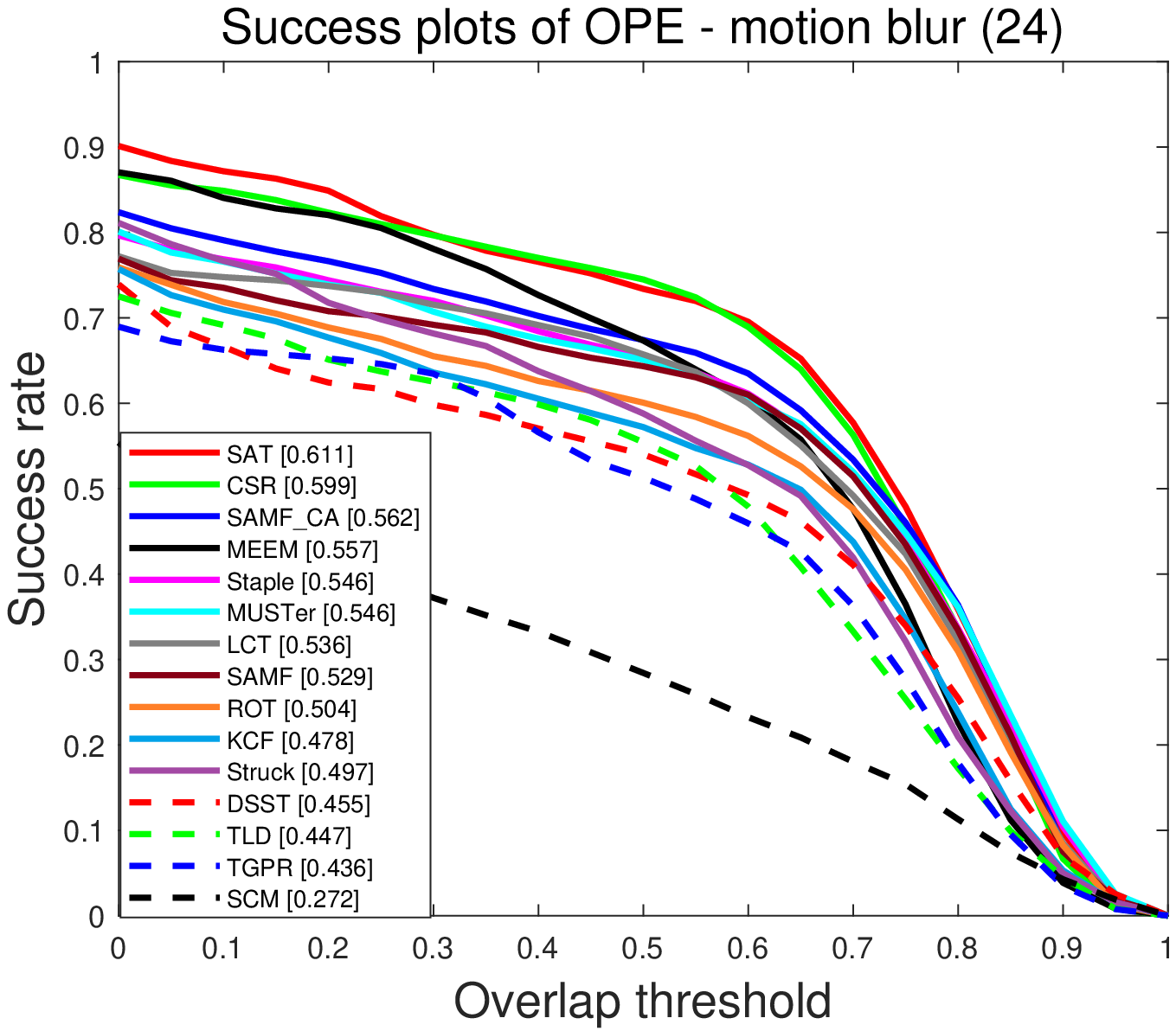}
\hspace{-0.5cm}
\includegraphics[width=6cm,height=5.2cm]{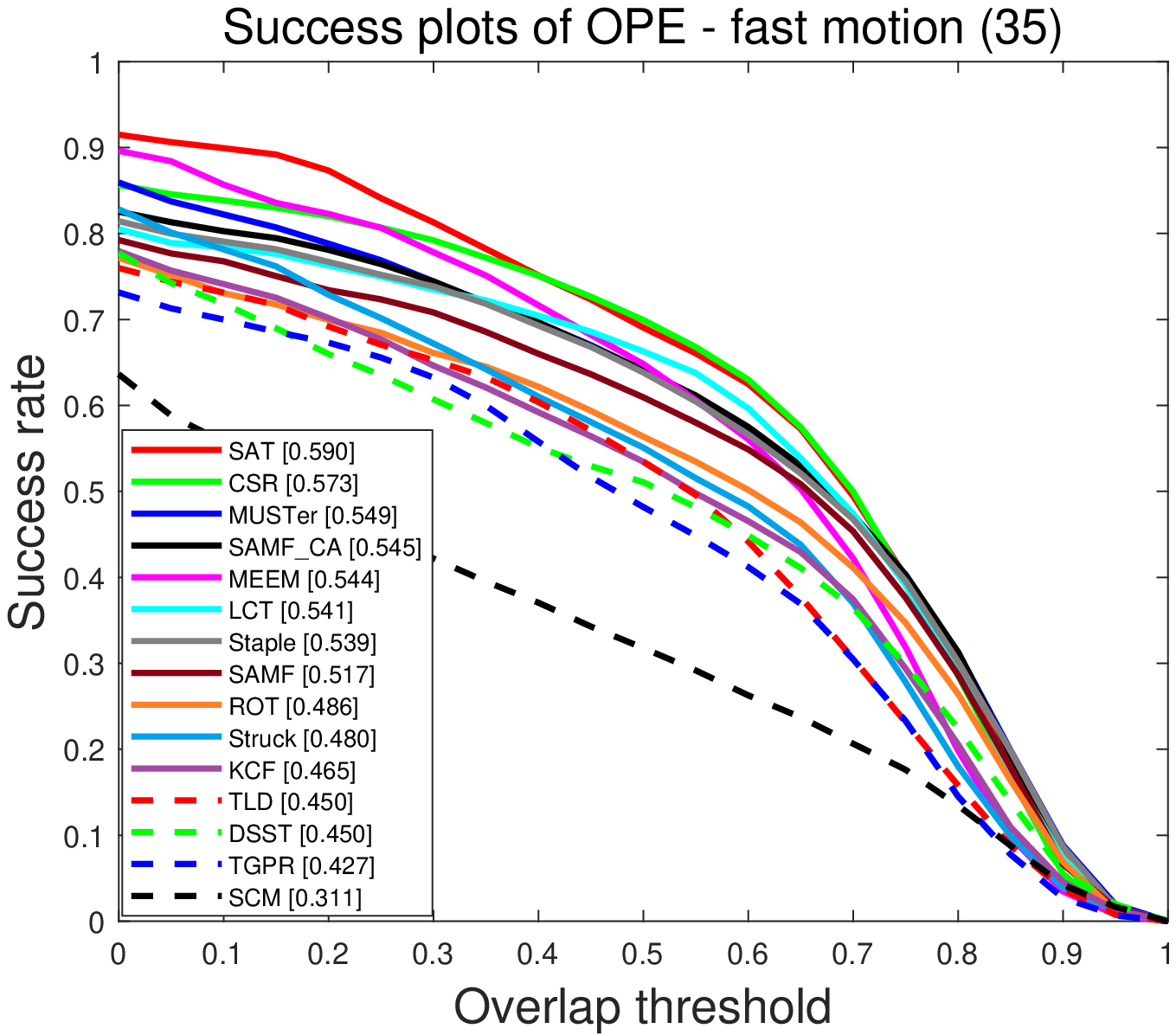} 
\hspace{-0.5cm}
\includegraphics[width=6cm,height=5.2cm]{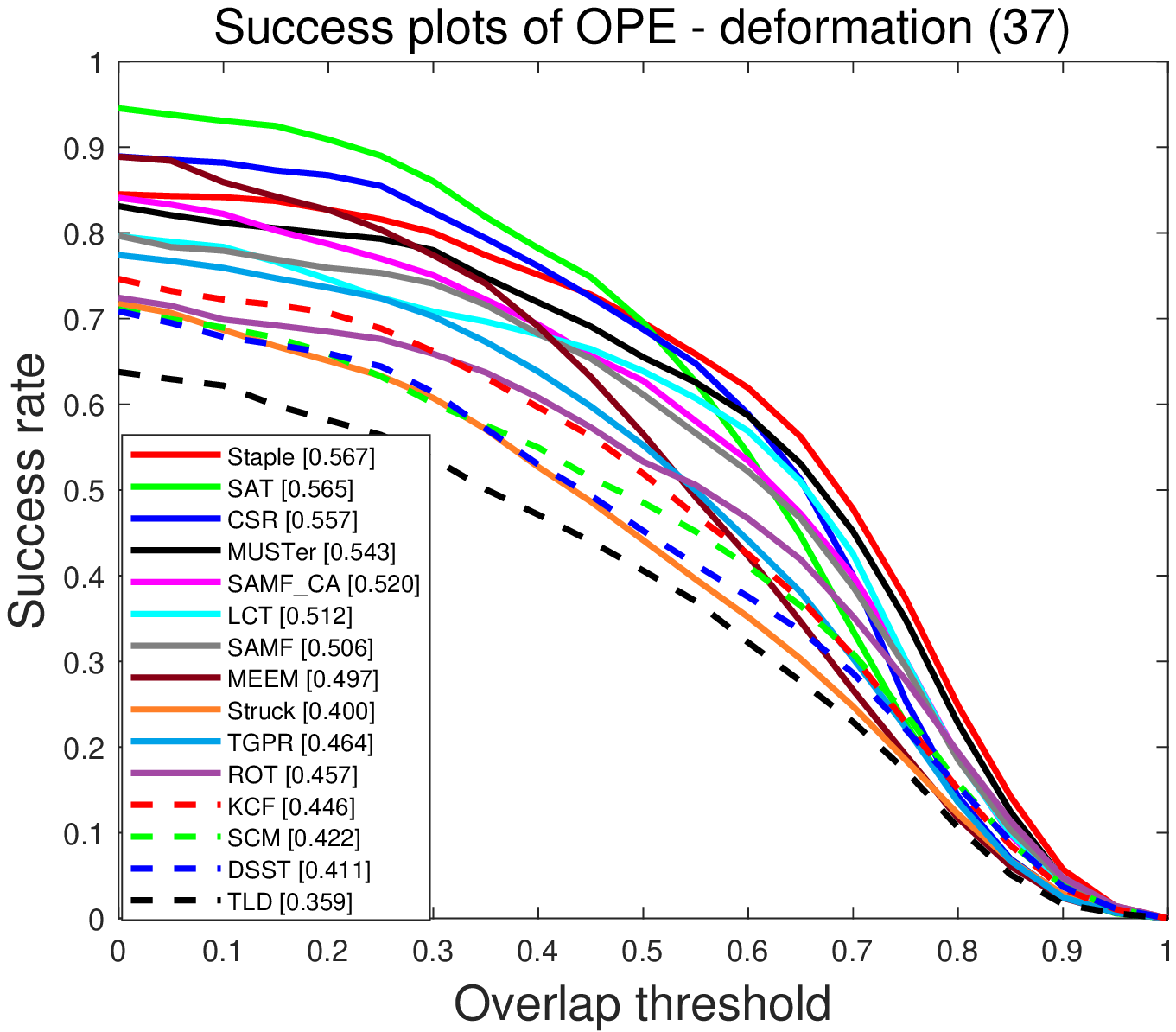} \\ \vspace{0.25cm}
\includegraphics[width=6cm,height=5.2cm]{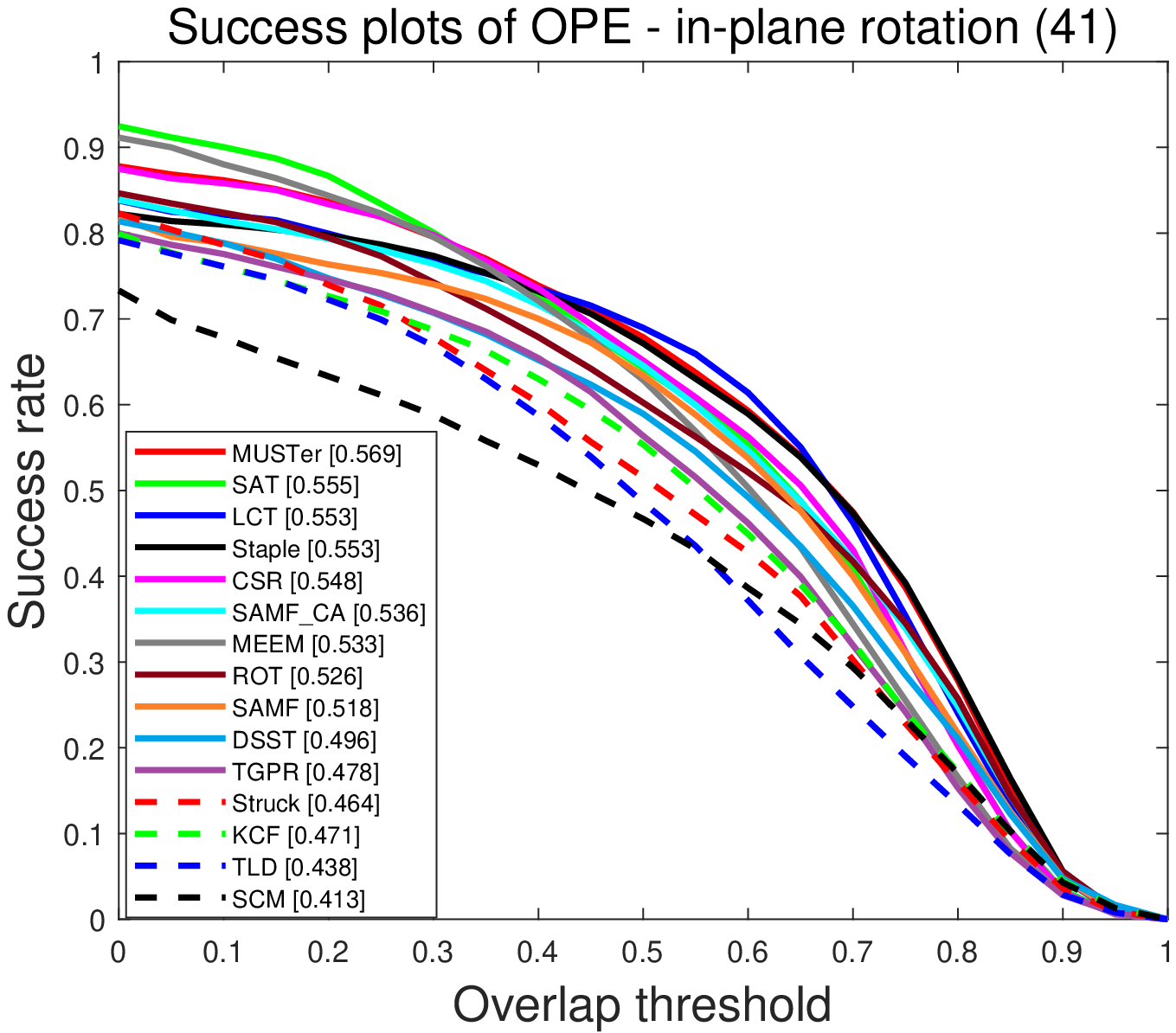} 
\hspace{-0.5cm}
\includegraphics[width=6cm,height=5.2cm]{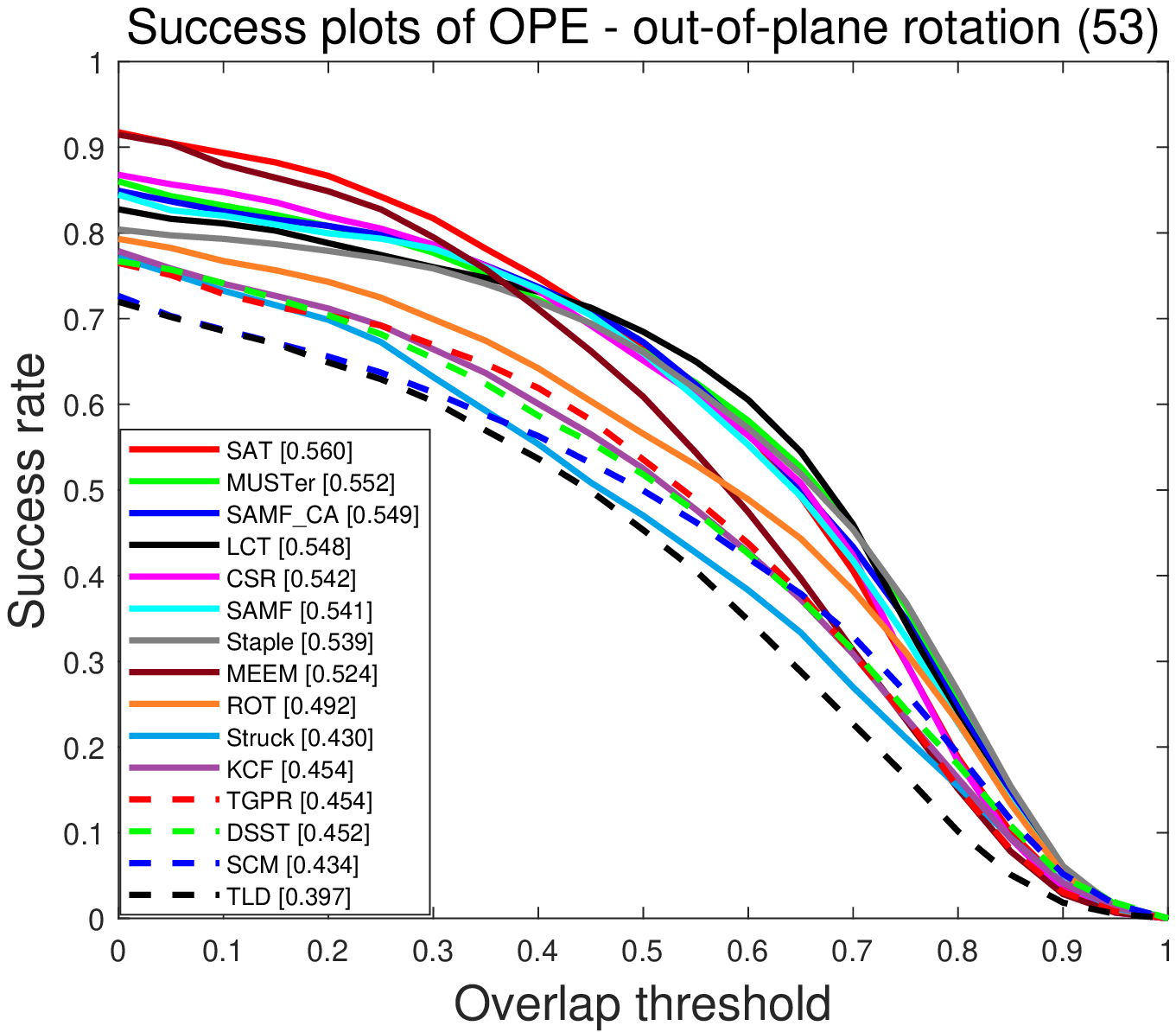}
\hspace{-0.5cm}
\includegraphics[width=6cm,height=5.2cm]{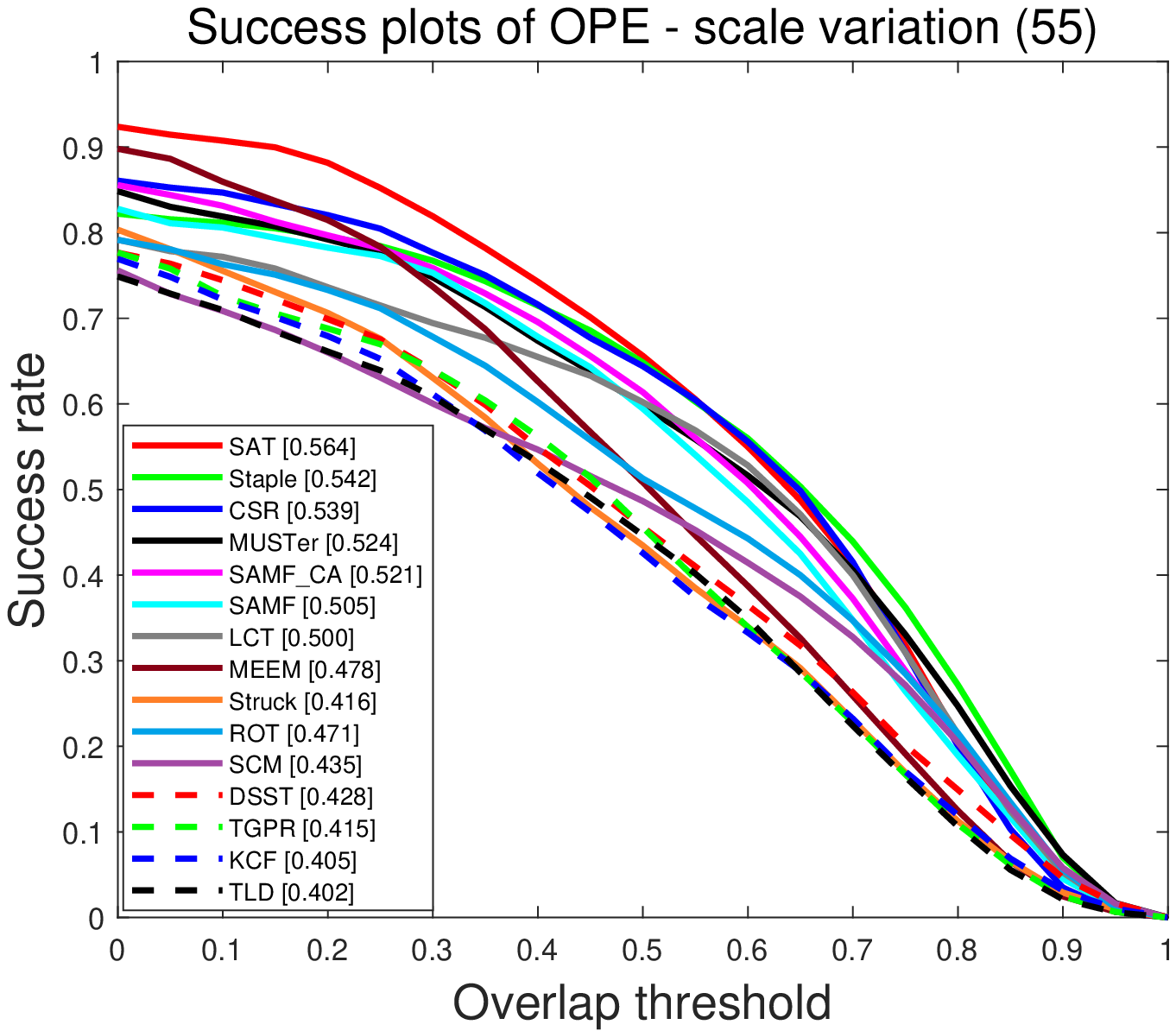} \\ \vspace{0.05cm}
\hspace{-0.5cm}
\hspace{2.5cm}
\vspace{-0.3cm}
\caption{The success plots of the videos with different attributes. The number in the title indicates the index of corresponding sequence. Best viewed in color.}
\label{fig:10}
\end{figure*}

\begin{figure*}[!t]
\vspace{0.5cm}
\centering
\includegraphics[width=6cm,height=5.2cm]{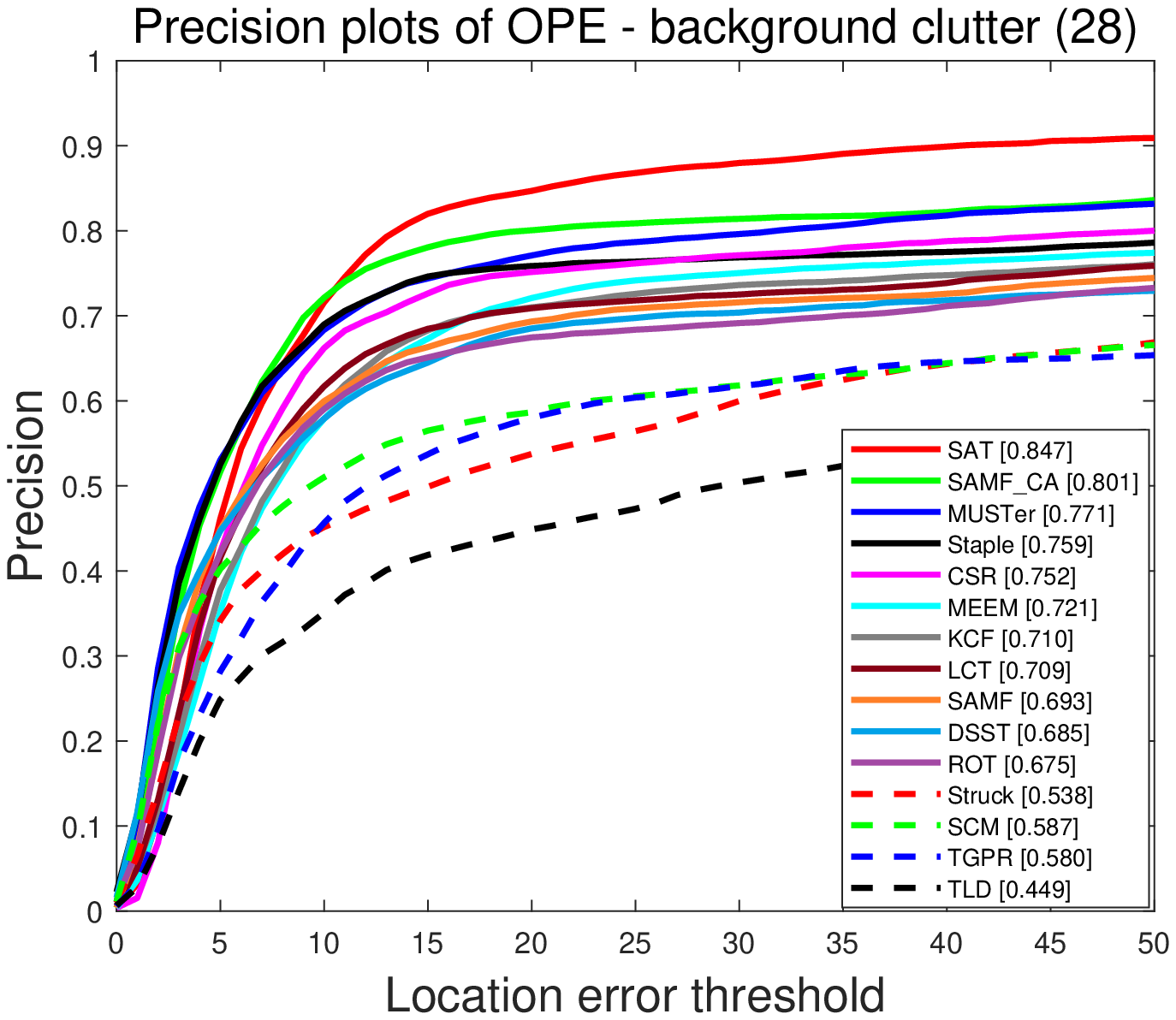}
\hspace{-0.5cm}
\includegraphics[width=6cm,height=5.2cm]{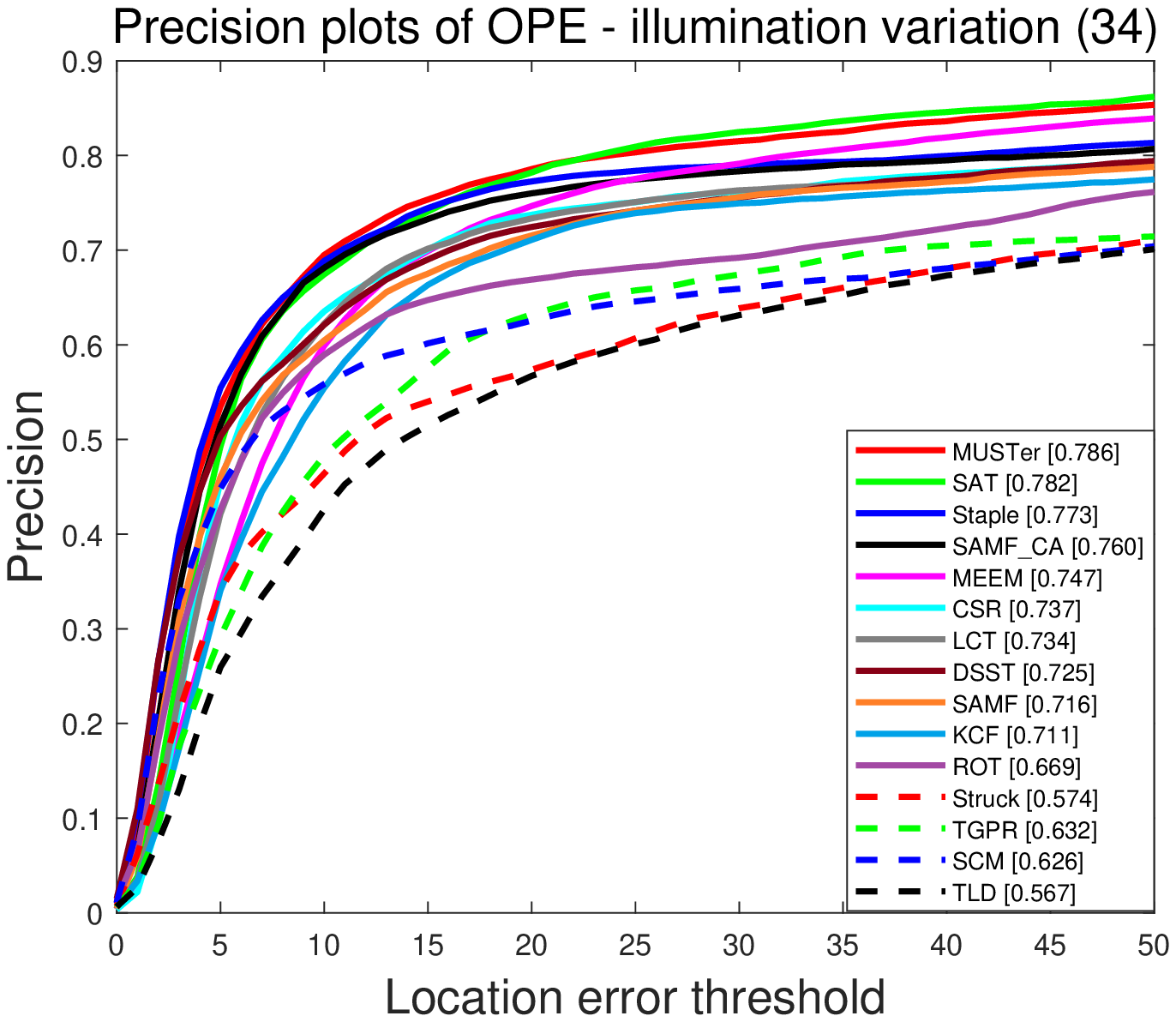}
\hspace{-0.5cm}
\includegraphics[width=6cm,height=5.2cm]{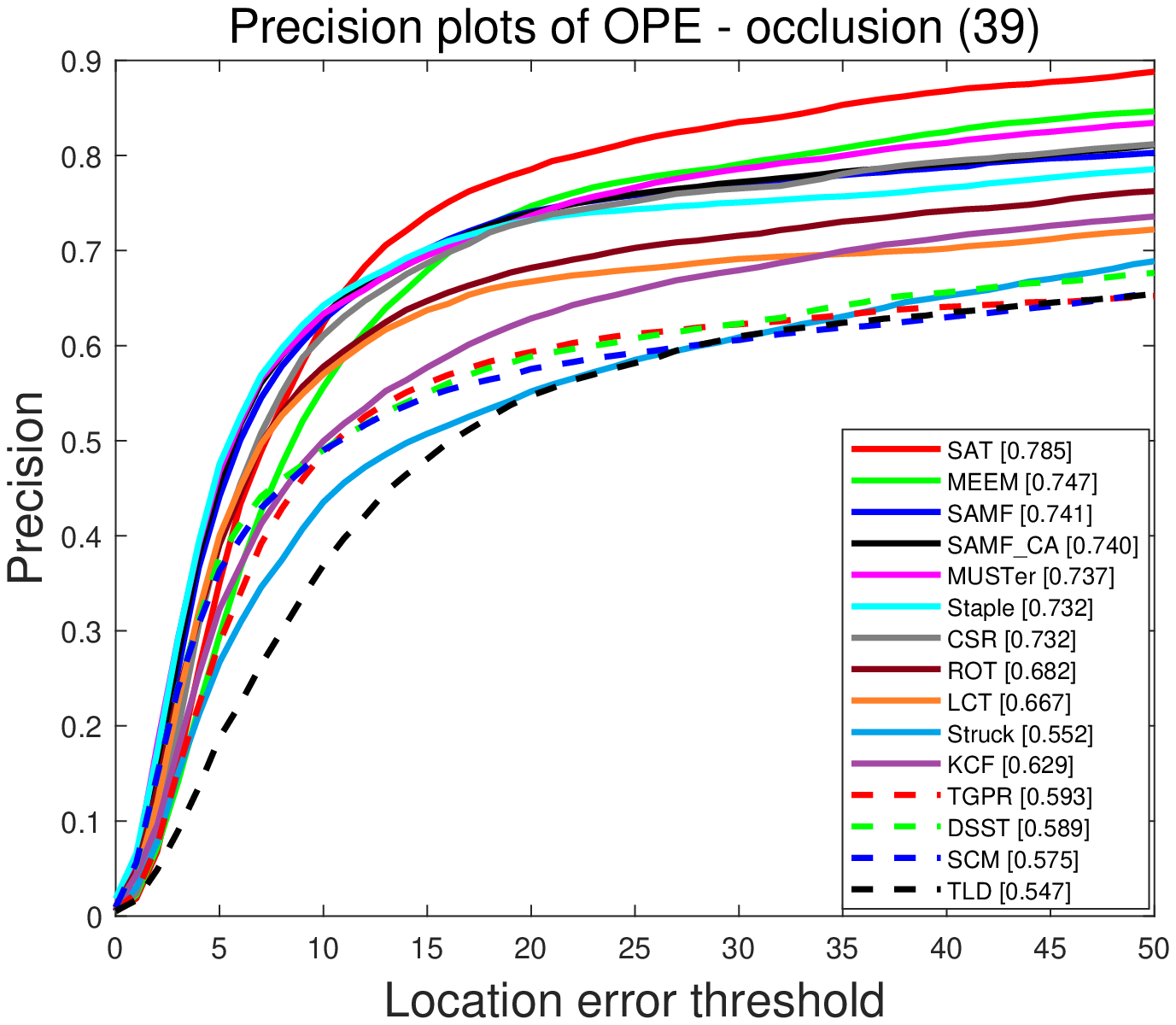}\\ \vspace{0.25cm}
\includegraphics[width=6cm,height=5.2cm]{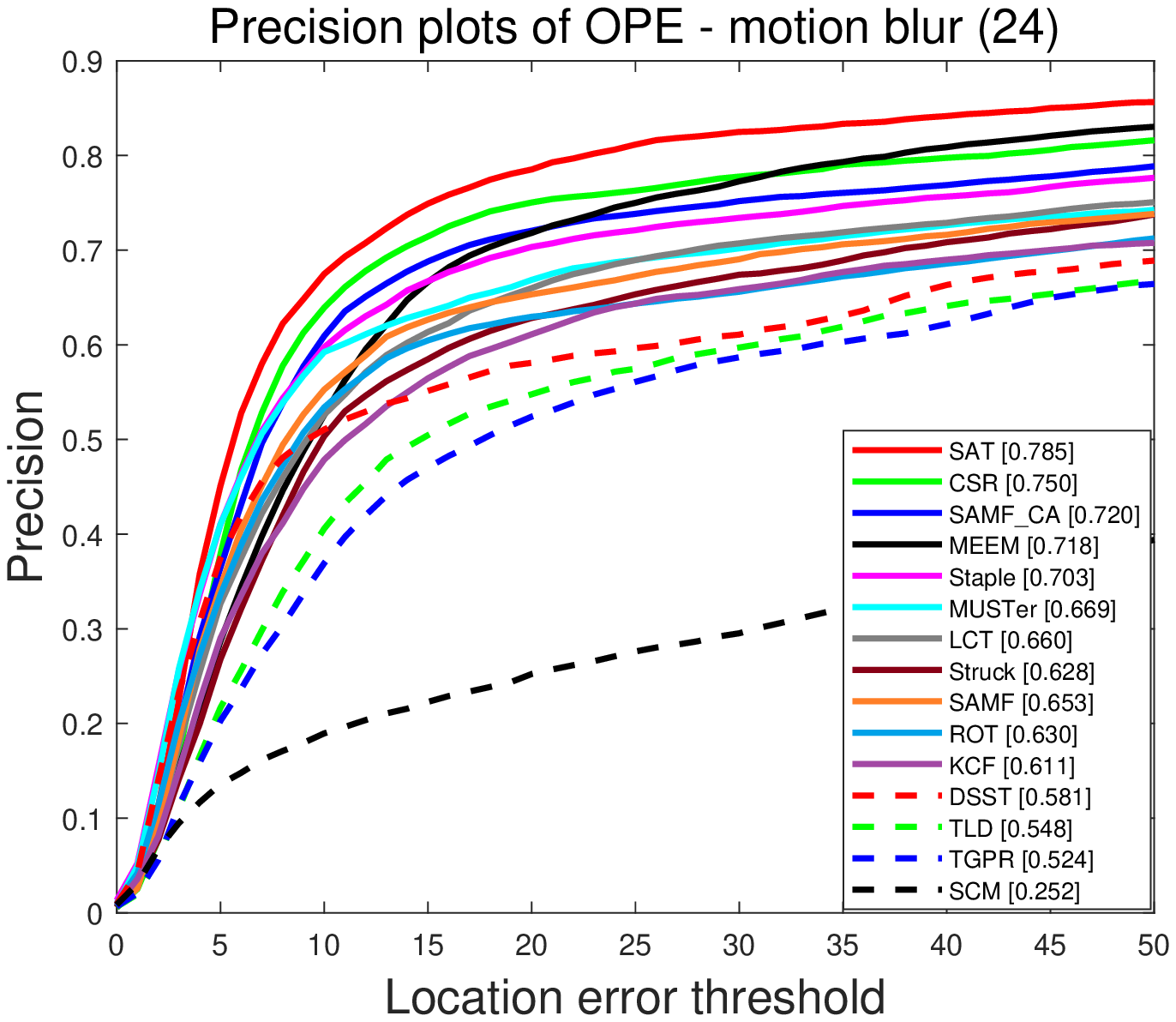}
\hspace{-0.5cm}
\includegraphics[width=6cm,height=5.2cm]{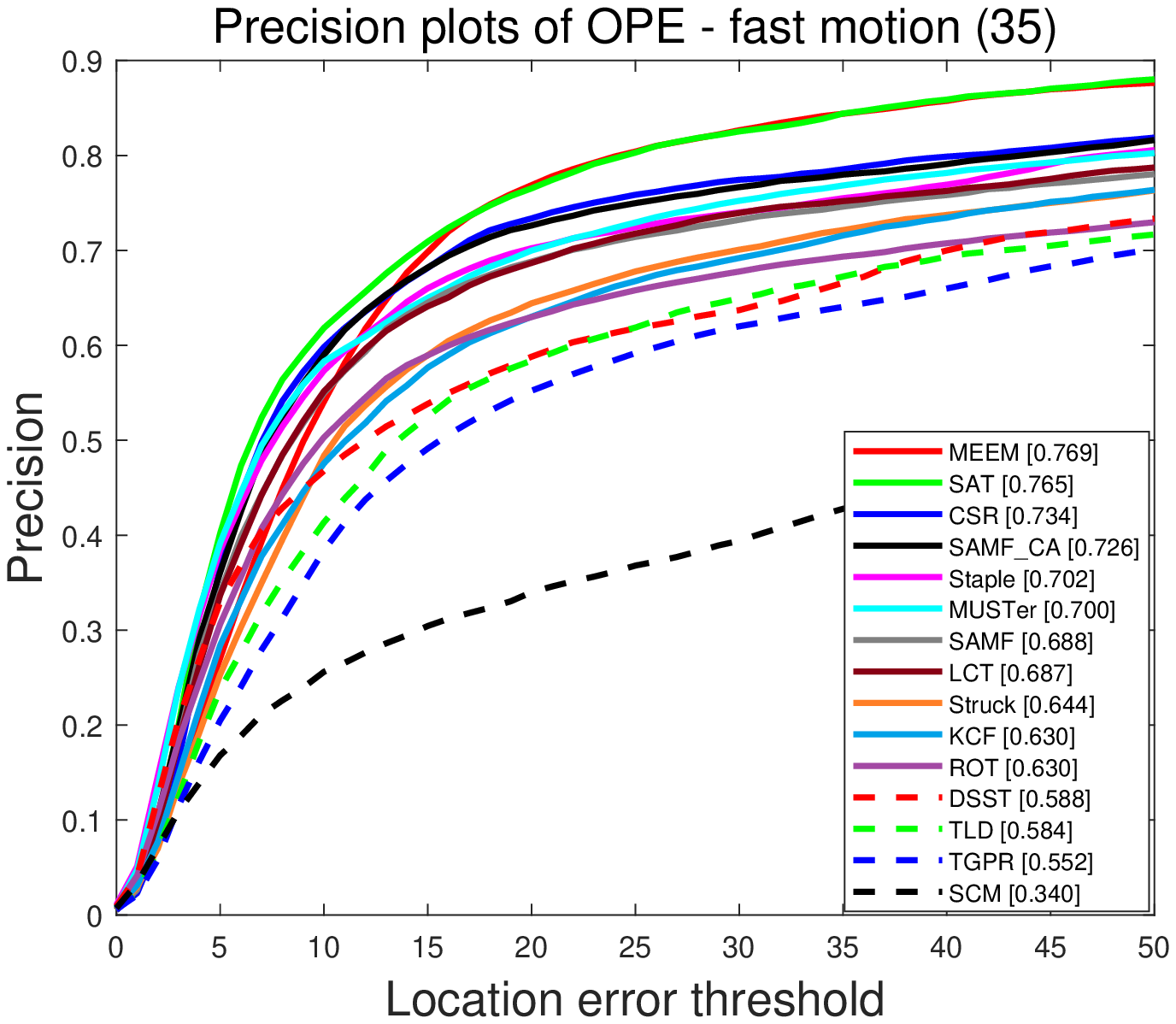}
\hspace{-0.5cm}
\includegraphics[width=6cm,height=5.2cm]{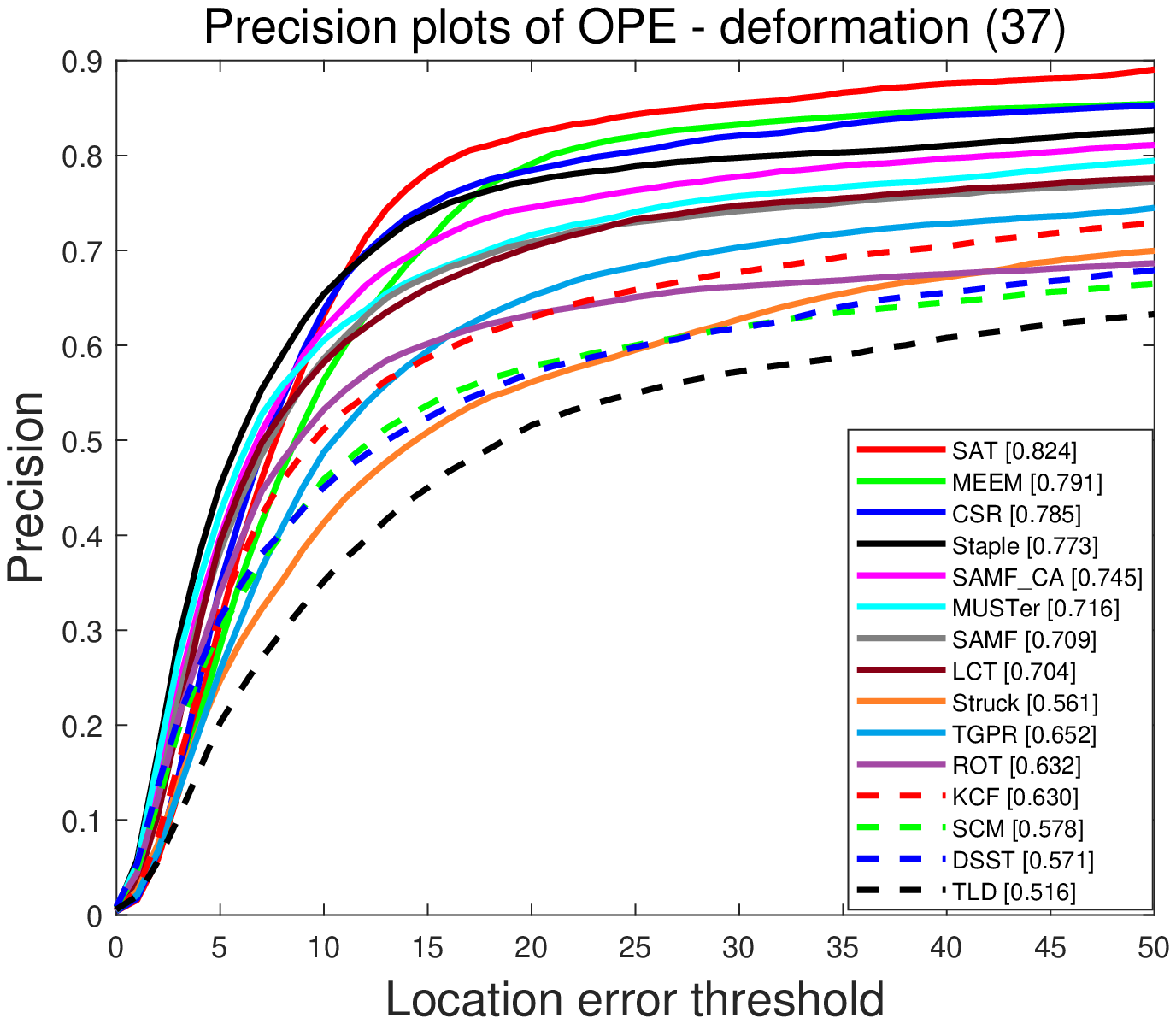}\\ \vspace{0.25cm}
\includegraphics[width=6cm,height=5.2cm]{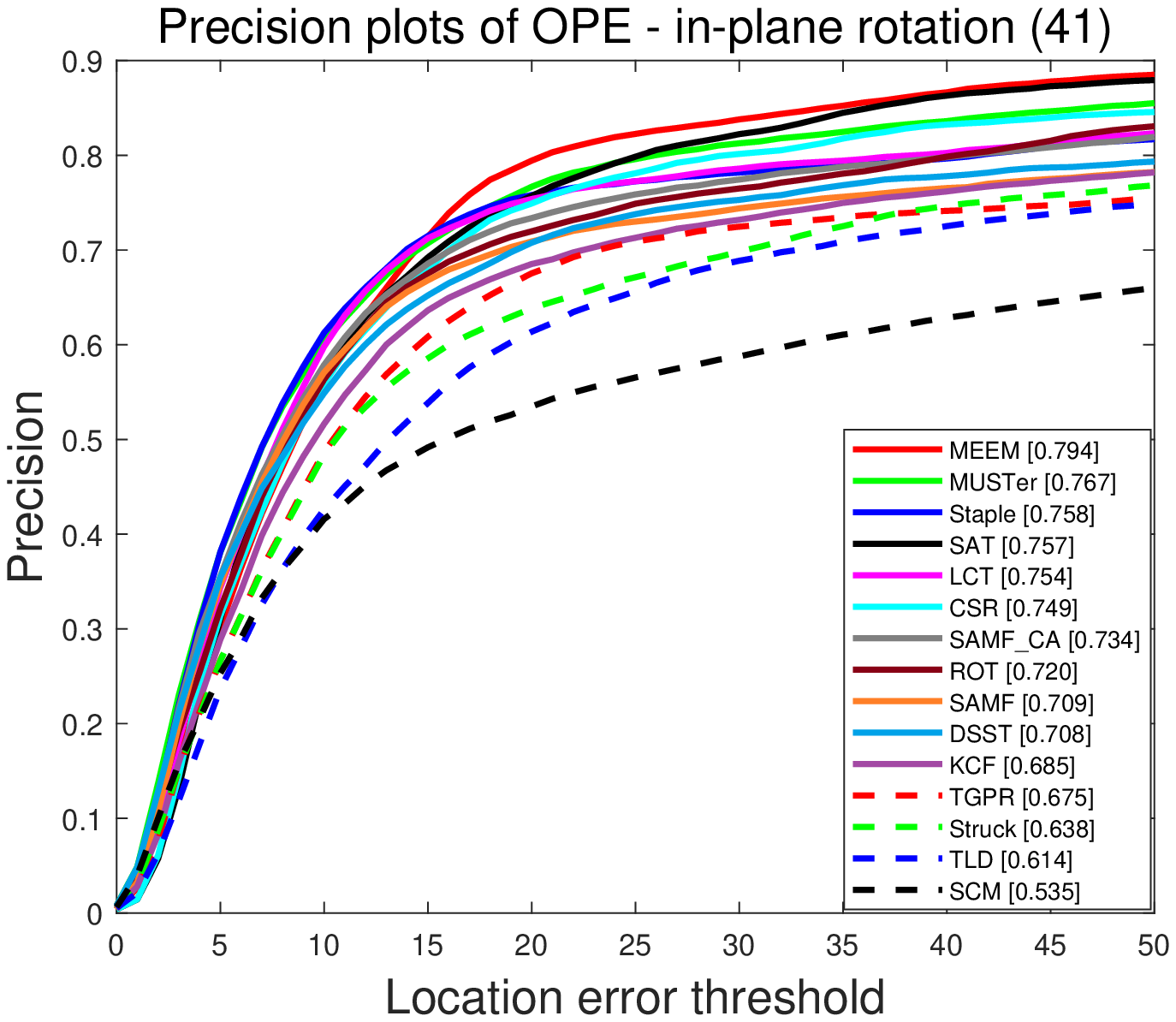}
\hspace{-0.5cm}
\includegraphics[width=6cm,height=5.2cm]{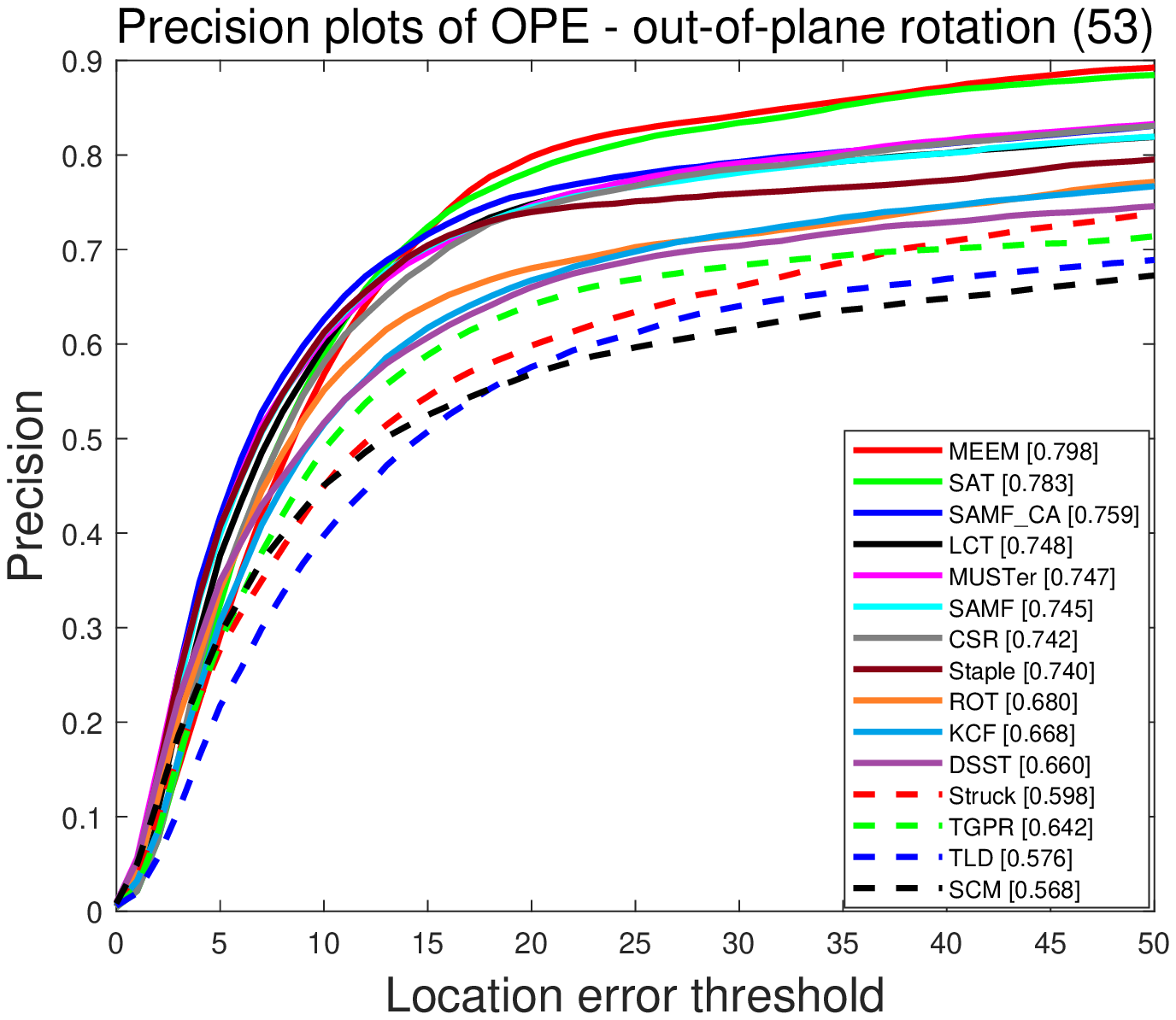}
\hspace{-0.5cm}
\includegraphics[width=6cm,height=5.2cm]{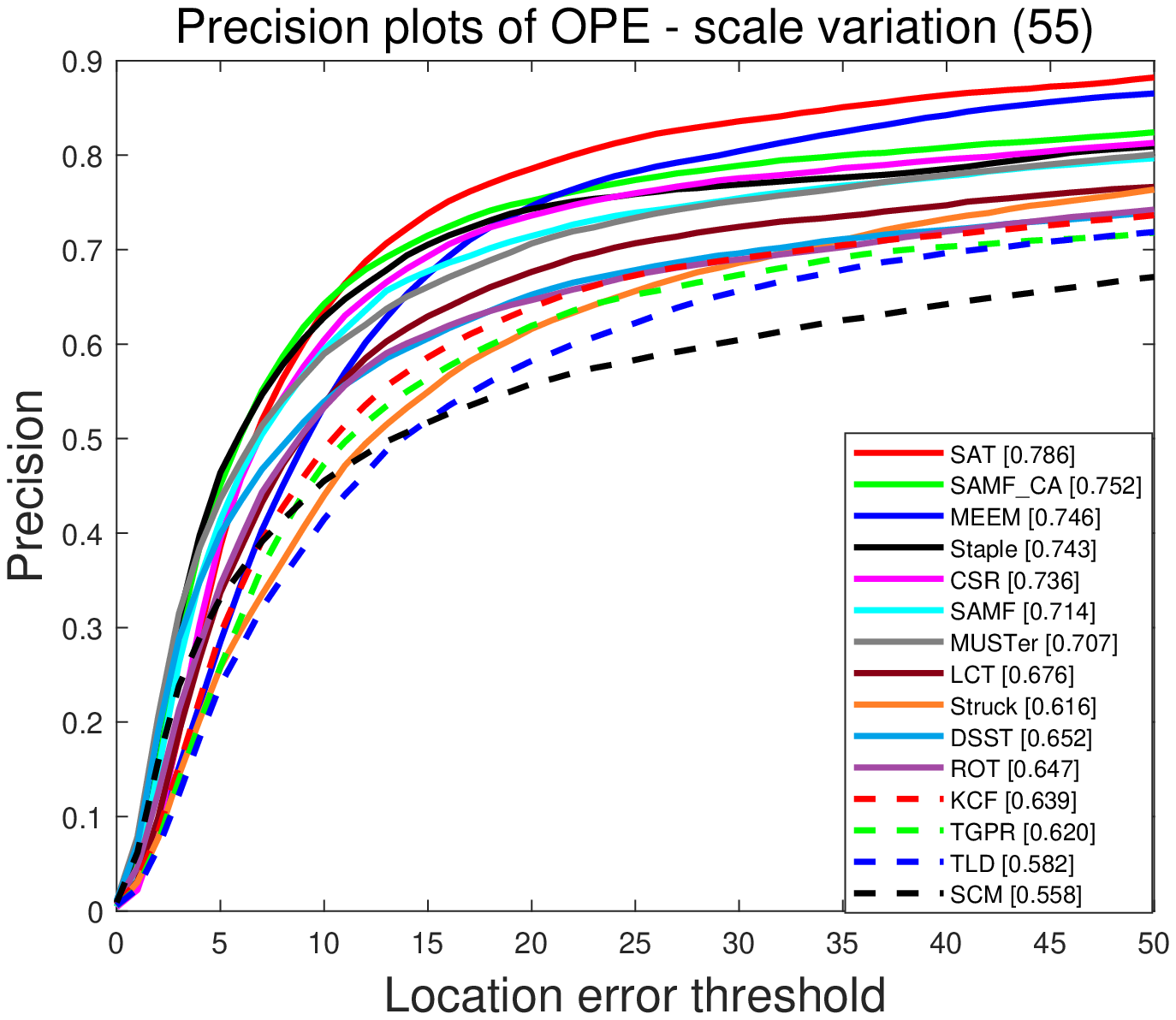} \\ \vspace{0.25cm}
\hspace{-0.5cm}
\hspace{2.5cm}
\vspace{-0.3cm}
\caption{The precision plots of the videos with different attributes. The number in the title indicates the index of corresponding sequence. Best viewed in color.}
\label{fig:11}
\end{figure*}

\subsection{Quantitative Comparison}
\label{ssec:4.3}
Figure 5 illustrates the overall performance of the 13 conventional trackers on OTB-100 in terms of success and precision plots. Among all the compared trackers, the proposed SAT tracker obtains the best performance, which achieves a 0.607 AUC score and a 0.837 distance precision rate at the threshold of 20 pixels. SAMF-CA is the baseline tracker for the proposed tracker, while SAT improves the tracking performance by 3.3 percent and 4.6 percent in success plot and precision plot respectively. In addition, the proposed tracker also outperforms the other state-of-the-art or milestone trackers by a distinct margin. 

To further evaluate the proposed tracker, we implement DeepSAT tracker with CNN  features. We conduct the same experiments with eight popular CNN based trackers including CNN-SVM \cite{d32}, CNT \cite{d33}, HCF \cite{d21}, CFNet \cite{d34}, Siamese \cite{d35}, HDT \cite{d19}, DeepSRDCF \cite{d36} SRDCFdecon \cite{d37}. The one-pass-evaluation of the two metric is shown in Table I. It can be seen that the DeepSAT tracker performs favorably comparing against other CNN-based trackers, which further demonstrates our tracker is effective and promising.

\begin{figure*}[!t]
\vspace{-0.5cm}
\centering{
\subfloat[Human3 and Jogging1 with Occlusion and Out-of-View]{
\includegraphics[width=16cm,height=4.cm]{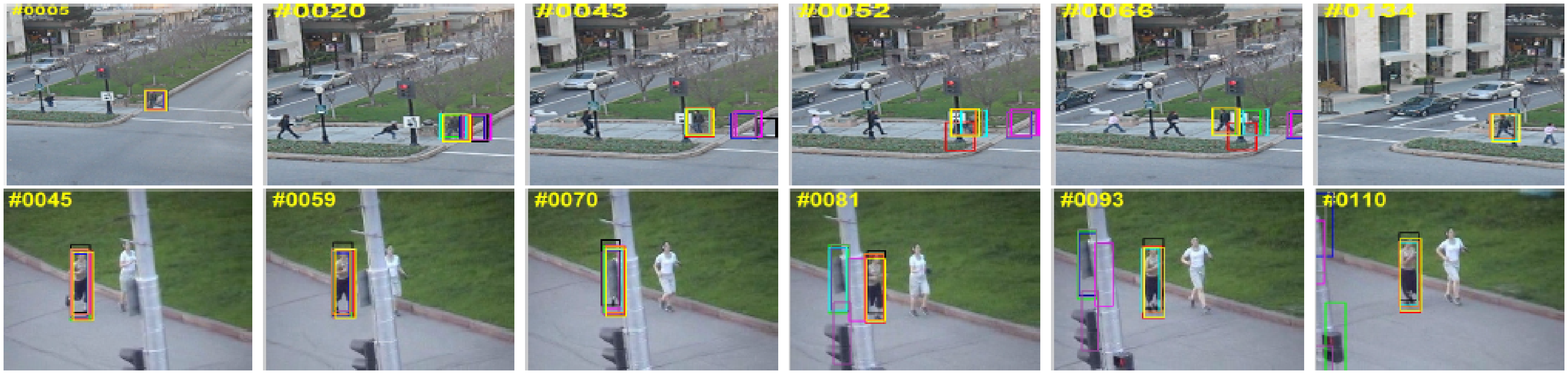}
} \\ \vspace{-0.4cm}
\subfloat[Clifbar, Bolt2 and Football with Background Clutters and Distractors]{
\includegraphics[width=16cm,height=5.6cm]{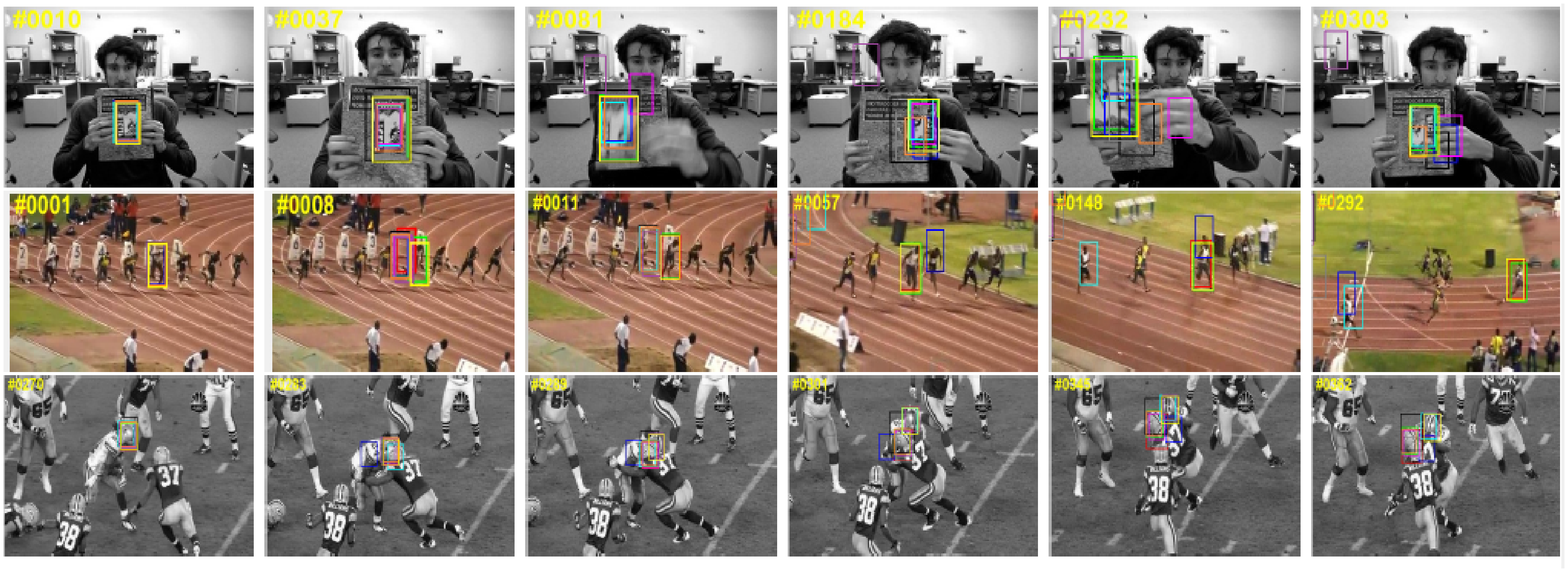}
} \\ \vspace{-0.4cm}
\subfloat[Couple and Shaking with Abrupt Motion and Illumination variation]{
\includegraphics[width=16cm,height=4.cm]{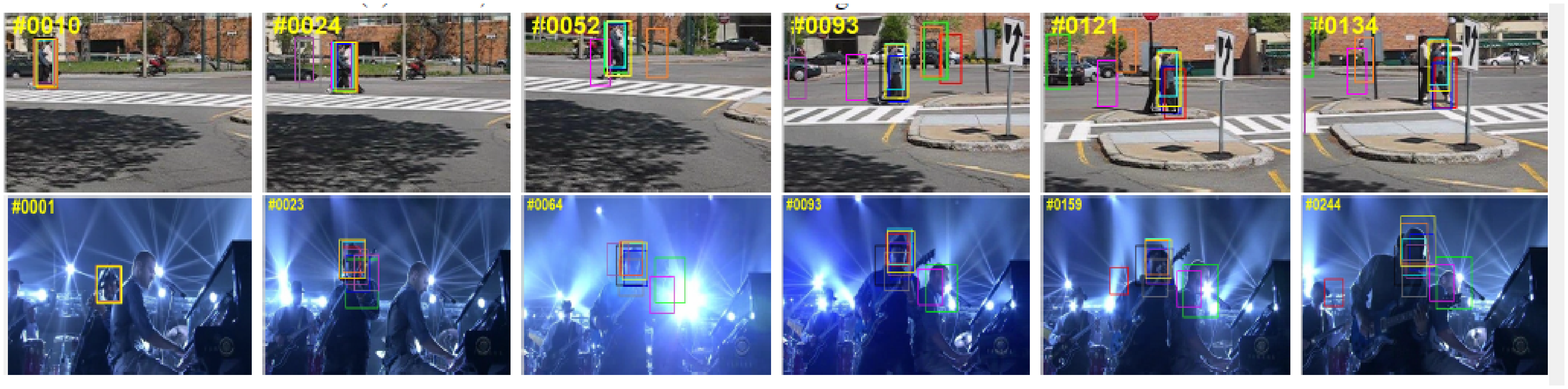}
} \\ \vspace{-0.4cm}
\subfloat[Diving, Skating and Motorrolling with Rotation and Deformation]{
\includegraphics[width=16cm,height=5.6cm]{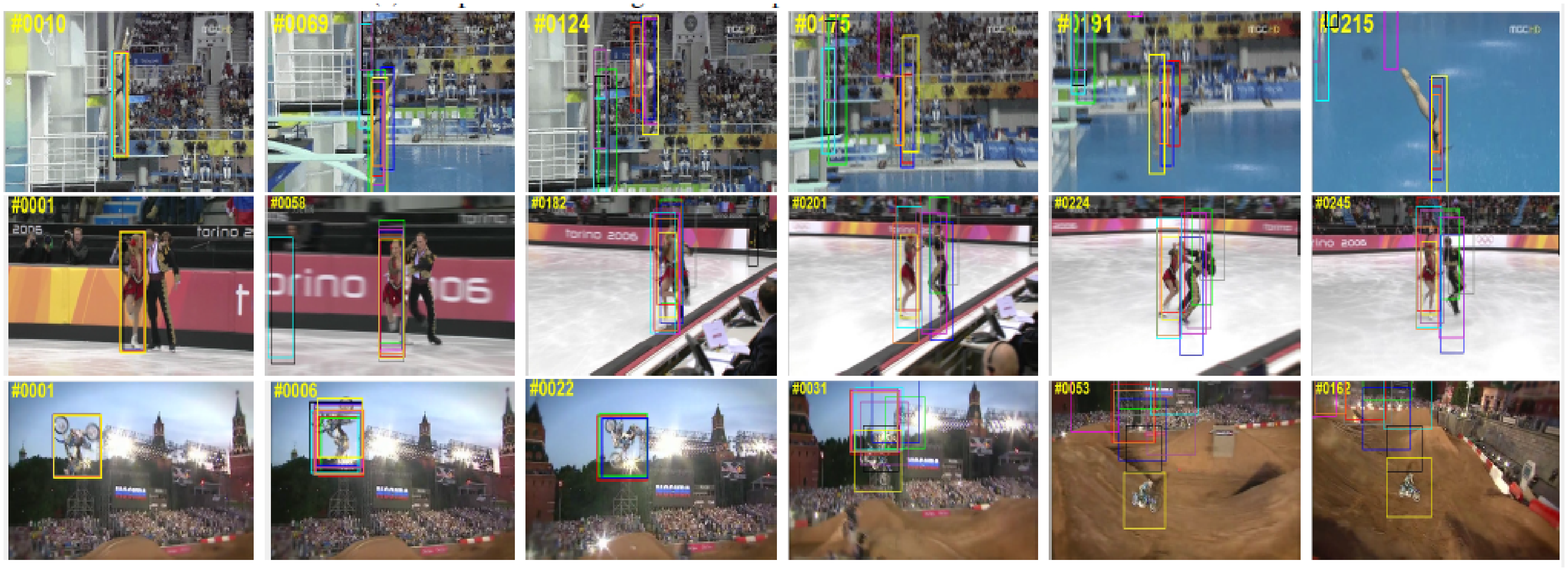}
} \\ \vspace{-0.4cm}
\subfloat[Human9 and Dog1 with Scale Variation]{
\includegraphics[width=16cm,height=4.cm]{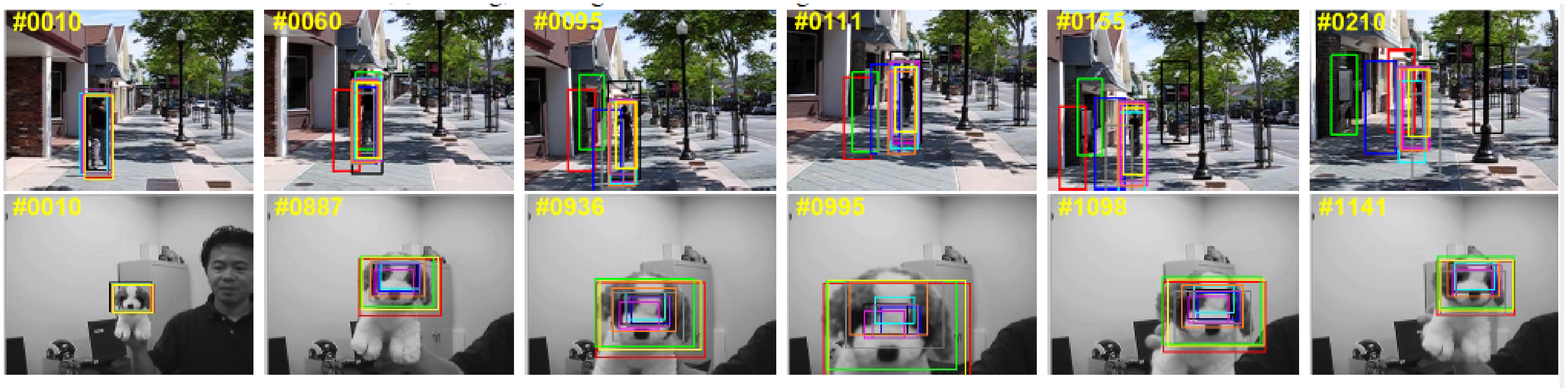}
}
}
\\ \vspace{-0.05cm}
\includegraphics[width=16cm,height=0.7cm]{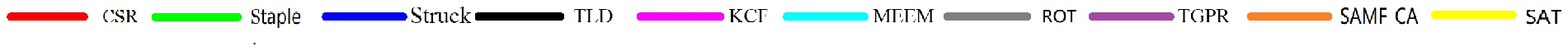}\\ \vspace{-0.5cm}
\caption{Representative tracking results on some challenging sequences. Best viewed in color.}
\label{fig:12}
\end{figure*}

Since the target undergoes different challenging attributes during tracking process, we note that it's crucial to investigate the tracking performance on these factors. Figure 6 and Figure 7 illustrate the success plot and precision plot on nine different factors which we have mainly discussed in this paper. We categorize these factors into two classes: the external interference (BC, IV, OCC and MB) and internal target appearance change (FM, DEF, IPR, OPR and SV). In most cases, the SAT tracker ranks top 2 among 15 trackers out of both the external and internal attributes synchronously. For external interference, our SAT tracker benefit from the learning of context patches, which could be aware of the potential distraction and  holistic appearance change in advance. On the other hand, the performance gain for internal target appearance change could be largely attributed to the reliable information learning in filter designment, which could train the filter with more accurate features compared with the conventional CF trackers. To sum up, by jointly considering the discrimination and reliability information in filter training stage, our SAT tracker performs more robust to the above challenging attributes as shown in Figure 6 and Figure 7.

\subsection{Qualitative Comparison}
\label{ssec:4.5}

To better visualize the tracking performance of SAT tracker, we provide a qualitative comparison of our approach with nine state-of-the-art trackers in Figure 7. Several video sequences are selected from OTB100 which contain various of challenging attributes to present the tracking performance among different trackers.

\subsubsection{Occlusion and Out of View}
\label{subsec:4.5.1}
During tracking, the target is often partially or fully occluded by other object, which will destroy the holistic appearance of the target. In other cases, some portion of the target may leave the view. Both of these attributes may lead to the model drift if the tracker is not robust enough. 
We evaluate the tracking algorithms on sequence Human3 and Jogging1 in Figure 8(a) that the people are occluded by   telegraph pole within some time during tracking. Unfortunately, most of the traditional CF trackers fail to track the target stably due to the corrupted training samples by the occluding objects. While our SAT tracker could address these issues, since we incorporate the surrounding context patches into training stage, which could detect the potential distractors in advance. In addition, the high confidence updating strategy could further guarantee the purity of tracking template.

\subsubsection{Background Clutter and Distractors}
\label{subsec:4.5.2}

The target in Sequence ClifBar has similar surrounding texture in background, it's hard to separate the foreground with background by extracting the positive samples' feature. Therefore, TGPR, KCF and ROT could hardly locate the target precisely without exploiting the background information, when the target undergoes background clutter. Moreover, even TLD is equipped with the re-detection module, it fails to recover from the previous drift due to its less-discriminative feature. Figure 8(b) also illustrates another two sequences, where there are similar objects appear in the screenshots. Most of the trackers fail to distinguish the target from the distractors and drift gradually (see the tracking result of MEEM at frame 148 in Bolt2 and the result of CSR at frame 345 on Football1).
Only our tracker successfully locates the people throughout the entire video in both videos. It could be attributed to the discrimination and reliability information adopted in training stage, which guarantee the filter focus on the reliable feature and suppress the potential distractions and false positives.   

\subsubsection{Abrupt Motion and Illumination Variation}
\label{subsec:4.5.3}

Figure 8(c) shows the screenshots of the tracking results in 2 challenging sequences where the object undergoes
abrupt motion and illumination variation. ROT, TGPR and Staple undergoes serve drift when the camera shakes. SAMF-CA, CSR and KCF fail to detect the target when the target moves abruptly at frame 93 in Couple sequence. TLD, MEEM and our SAT tracker ultimately complete the challenging task. This is largely due to the fact that adding context patches allows for a larger search region and make the tracker insensitive to abrupt motion. In sequence Shaking, CSR and Staple lose the target when illumination variation occurs, since the color histogram adopted in their tracker are notoriously sensitive to the light change. Nevertheless, our SAT tracker combine the color cues with the surrounding context cues in a complementary manner, which could ensure the filter concentrate on the reliable part as well as not interfered by external factors. In addition, the high confidence updating scheme could  monitor the tracking condition frame by frame, forecast the potential tracking failure and guarantee the purity of the learned filter.

\subsubsection{Rotation and Deformation}
\label{subsec:4.5.4}

Figure 8(d) shows the screenshots of the tracking results in 3 challenging sequences where the object undergoes rotation and large shape deformation. In Diving and Skating sequence, CSR, SAMF-CA and our SAT tracker ultimately complete the challenging task, while other tracker either lose the target or drift on other distractors. In MotorRolling sequence, all the other trackers fail to track the motor due to the rapid rotation and deformation. In contrast, SAT tracker locates the target precisely through the entire sequence. We owe the superior performance to the effectiveness of color-based reliability learning, since the color statistics cope well with the variation in shape and rotation. 

\subsubsection{Scale Variation}
\label{subsec:4.5.5}
During tracking, the scale of the target often varies in successive frames as the target's movement. Hence, a tracker is required to estimate the scale as accurately as possible. Figure 8(e) shows the tracking results on human9 and Dog1 sequences with large scale changes. Some of the trackers could not tackle this issue and gradually drift due to the  error accumulating even though they have equipped with the scale adaption scheme (See the tracking results of CSR, Staple and Struck tracker at Human9 sequence). While the proposed tracker could deal with the above challenges and performs better than the other trackers achieving a long-term stable tracking with the employment of effective searching strategy and anti-drift filter learning mechanism.

\section{Conclusion}
\label{sec:5}

In this paper, we propose a generic framework for correlation filter (CF) based tracker, which jointly consider the discrimination and reliability information in filter learning stage. Context patches are employed into filter training stage to better distinguish the target from backgrounds. Furthermore, a color-based reliable mask is learned each frame to encourage the filter focus on more reliable regions suitable for tracking. Compared to the existing CF-based trackers, the proposed tracker handles not only the tracking challenges caused by external attributes but also the issues with target internal appearance change. Numerous experiments demonstrate the effectiveness and robustness of the proposed tracker (SAT and DeepSAT) against other relevant state-of-the-art methods.

\vspace{0.cm}

\section*{Acknowledgment}
The authors are grateful to the anonymous reviewers for their encouraging and insightful advice that lead to this improved version and clearer presentation of the technical content. This work is partially supported by the National Natural Science Foundation of China (NSFC) Grant 91438203 and 3172901. This work is also partially supported by Changjiang Scholars Programme (No. T2012122). All of the authors are with the Beijing Key Laboratory of Embedded Real-Time Information Processing Technology, School of Information and Electronics, Beijing Institute of Technology, Beijing 10081, China. The corresponding author is Chenwei Deng, with the corresponding e-mail: cwdeng@bit.edu.cn.


%
%
%
\vspace{0.3cm}
\section*{Apendix A}
This section provides the complete derivation of the solutions for equation 8 in the manuscript. As discussed in  the Section III-B, the augumented Lagrangian objective function $L({\hat{\bf w}}_{c},{\bf w},{\hat{\bf I}},\rho)$ is defined as :

\vspace{-0.3cm}
\begin{eqnarray}
L({\hat{\bf w}}_{c},{\bf w},{\hat{\bf I}},\rho) = \|{\bf B{\hat {\bf w}_c}-\hat{{\bf Y}}}\|_{2}^{2} + \lambda_{1}\|\hat{\bf w}_{r}\|_2^2  \nonumber\\
+ \hat{\bf I}^T(\overline{\hat{\bf w}_c}-\overline{\hat{\bf w}_r}) + \rho\|(\hat{\bf w}_c-\hat{\bf w}_r) \|_2^2
\end{eqnarray}

Where, ${\bf B}$ is the stacked feature matrix forming by the target patch and K context patches, ${\bf Y}$ denotes the new regression label corresponding to the target patch and image patches. Meanwhile, $\hat{\bf w}_{r}$ denotes the Hadamard product between the reliability mask and the base filter, ${\bf w}_{r}={\bf r}\odot{\bf w}$.
According to the property for Fourier Transform, $\hat{\bf w}_{r} = \sqrt{N}{\bf F}{\bf R}{\bf w}$. Here, {\bf F} is the orthonormal matrix of Fourier coefficient with the size $N\times N$ and ${\bf R}$ equals to $diag({\bf r})$. Afterwards, we would solve the Augmented Lagrangian objective function. We would like to decompose the overall problem down into four constitute parts for clearer representation and simplification.

\vspace{-0.2cm}
\begin{eqnarray}
\vspace{-0.2cm}
L_{1} &=& \|{\bf B{\hat {\bf w}_c}-\hat{\bf Y}}\|_{2}^{2}  \nonumber  \\
    &=& \hat{\bf w}_{c}^{H}{\bf B}^{H}{\bf B}\hat{\bf w}_{c}- \hat{\bf w}_{c}{\bf B}^{H}\hat{\bf Y}-{\bf B}\hat{\bf w}_{c}\hat{\bf Y}^{H}+\hat{\bf Y}^H\hat{\bf Y}  \nonumber 
\\
L_{2} &=&  \lambda_{1}\|\hat{\bf w}_{r}\|_2^2  =  \lambda_{1}N{\bf F}{\bf R}{\bf w}{\bf w}^{H}{\bf R}^{H}{\bf F}^{H}   = \lambda_{1}{\bf R}{\bf w}{\bf w}^{H}    \nonumber
\\
L_{3} &=& \hat{\bf I}^{T}\overline{\hat{\bf w}_{c}}-\hat{\bf I}^{T}\overline{\hat{\bf w}_{r}} \nonumber
\\
L_{4} &=&  \rho\|(\hat{\bf w}_c-\hat{\bf w}_r) \|_2^2 \nonumber  \\
  &=& \rho(\hat{\bf w}_{c}^{H}\hat{\bf w}_{c}-\hat{\bf w}_{c}\sqrt{N}\overline{{\bf F}{\bf R}{\bf w}}-\hat{\bf w}_{c}^{H}\sqrt{N}{\bf F}{\bf R}{\bf w}+ D{\bf R}{\bf w}{\bf w}^{H})  \nonumber
\end{eqnarray}

According to the nature of ADMM model, the objective function could be solved using a series of iterations as:

\vspace{-0.1cm}
\begin{equation}  
\left\{  
            \begin{array}{lr}  
            \hat{\bf w}_{c}^{i+1} = \arg\min\limits_{{\bf w}_c} L({\hat{\bf w}}_{c}^{i},{\bf w}^{i},\hat{\bf I}^{i},\rho^{i})  \nonumber \\
            {\bf w}^{i+1} =\arg\min\limits_{{\bf w}} L({\hat{\bf w}}_{c}^{i+1},{\bf w}^{i},{\hat{\bf I}^{i}},\rho^{i}) \\ 
            \hat{\bf I}^{i+1} = \hat{\bf I}^{i} + \rho^{i}(\hat{\bf w}_{c}^{i+1} - \hat{\bf w}_{r}^{i+1} ) \nonumber \\ 
            \rho^{i+1} =  \min(\rho_{max},\beta\rho^{i}) \nonumber 
            \end{array}  
\right.  
\end{equation}

It could be viewed from the above equation that, for each iteration, we would find the optimal value of one parameter by fixing the others. Besides that, the value of $\rho$ is increasing in each iteration to guarantee the convergence in the standard ADMM technique. We follow such regulation by setting a multiplier $\beta$ to 3. Traditionally, the ADMM stops when the residual term $\hat{\bf w}_{c}^{i+1} - \hat{\bf w}_{r}^{i+1}$ is small enough. We fix the iteration times as five since the residual error decreases largely after first five iteration after experimental validation.

Seen from the above analysis, we could obtain the optimization over $\hat{\bf w}_{c}$ by setting the complex gradient of the Augmented Lagrangian Function to zero.
\begin{eqnarray}
\triangledown_{\hat{\bf w}_{c}^{H}}L = 0 
\end{eqnarray} 

Therefore, we could decompose the complex gradient into four parts as: $\triangledown_{\hat{\bf w}_{c}^{H}}L_{1}+\triangledown_{\hat{\bf w}_{c}^{H}}L_{2}+\triangledown_{\hat{\bf w}_{c}^{H}}L_{3}+\triangledown_{\hat{\bf w}_{c}^{H}}L_{4} = 0 $.
With the defination of these components, we could solve the partial gradient successively:

\begin{eqnarray}
\frac{\partial L}{\partial{\hat{\bf w}_{c}^{H}}} = ({\bf B}^H{\bf B}\hat{\bf w}_{c})-{\bf B}^{H}\hat{Y}+\hat{\bf I}^{T}+ \rho\hat{\bf w}_c +\rho\sqrt{N}{\bf N}{\bf F}w 
\end{eqnarray} 

Thus, we could yield the optimal value of $\hat{\bf w}_c$ for $i^{th}$ iteration by identifying the left to the right as follow:  

\begin{eqnarray}
\vspace{-0.2cm}
\hat{\bf w}_{c} = ({\bf B}^H{\bf B}+\rho)^{-1}(\rho\sqrt{N}{\bf F}{\bf R}w + {\bf B}^{H}\hat{{\bf Y}} -\hat{\bf I}^{T})
\end{eqnarray} 

Here, ${\bf B} = \left [ {\bf A_0}, \sqrt{\lambda_2}{\bf A_1},...,\sqrt{\lambda_2}{\bf A_k }\right ]^T$, denoting the stacked matrix for the feature matrix of target patch ${\bf A_{0}}$ and the ones of surrounding patches ${\bf A_1}$ to ${\bf A_k}$. Meanwhile, ${\bf Y}$ indicates their corresponding regression label ${\bf Y} = \left [{\bf y} , {\bf 0} ,..., {\bf 0} \right ]^T $, we manually set the label of the surrounding context to zero, treating them as negative samples. With that assumption, the SAT tracker would be able to detect the surrounding distractors by learning their features in advance.

Since, we have ${\bf B}^H{\bf Y} = {\bf A}_0^H\hat{\bf y}$ and ${\bf B}^H{\bf B} = {\bf A}_0^H{\bf A}_0+\lambda_2{\bf A}_1^H{\bf A}_1+...+\lambda_2{\bf A}_k^H{\bf A}_k = {\bf A}_0^H{\bf A}_0 + \lambda_2\sum_{i=1}^{k}{\bf A}_i^H{\bf A}_i$. Recalling the property of circulant matrix that, ${\bf X} = {\bf F}diag({\bf {\hat x}}){\bf F}^H$ and ${\bf X}^H = {\bf F}diag({\bf {\hat x}^{*}}){\bf F}^H$.  The above variable could be represented in element-wise in Fourier domain:

\vspace{-0.2cm}
\begin{eqnarray}
\vspace{0.1cm}
{\bf B}^H{\bf B}  &=& {\bf F} diag(\hat{a}^{*}_0\odot\hat{a}_0) {\bf F}^{H}+ \lambda_2\sum_{i=1}^{k} {\bf F} diag(\hat{a}^{*}_i\odot\hat{a}_i) {\bf F}^{H} \nonumber \\
&=&  {\bf F} diag(\hat{a}^{*}_0\odot\hat{a}_0+\lambda_2\sum_{i=1}^{k}\hat{a}^{*}_i\odot\hat{a}_i) {\bf F}^{H}\\
{\bf B}^H{\bf Y} &=& {\bf A}_0^H\hat{\bf y} = {\bf F} diag(\hat{a}^{*}_0\odot\hat{y}){\bf F}^{H} 
\end{eqnarray} 

$\hat{a}_0$ and $\hat{a}_i$ indicate the feature vector for target patch and context patches, in conjunction with the circulant feature matrix ${\bf A}_0$ and ${\bf A}_i$, respectively.

Therefore, by substituting the equation X with equation X, we obtain the element-wise closed-form solution for $\hat{\bf w}_c$:

\begin{eqnarray}
\hat{\bf w}_c = \frac{\hat{a}^{*}_0\odot\hat{y}+\rho\hat{\bf w}_r-\hat{\bf I}^T}{\hat{a}^{*}_0\odot\hat{a}_0+\lambda_2\sum_{i=1}^{k} \hat{a}^{*}_i\odot\hat{a}_i +\rho}
\end{eqnarray} 

Similarly, we could acquire the optimal ${\bf w}$ by setting the complex gradient of the Argmented Lagrangian function to zero, that is $\triangledown_{{\bf w}^{H}}L = 0 $.

\begin{eqnarray}
\triangledown_{{\bf w}^{H}}L_{1}+\triangledown_{{\bf w}^{H}}L_{2}+\triangledown_{{\bf w}^{H}}L_{3}+\triangledown_{{\bf w}^{H}}L_{4} = 0 
\end{eqnarray} 

By solving the formulation, we obtain the following result:
\begin{eqnarray}
\frac{\partial L}{\partial{{\bf w}^{H}}} = \lambda_1N{\bf R}{\bf w}+\rho N{\bf R}{\bf w}-\rho\hat{\bf w}_c\sqrt{N}{\bf R}{\bf F}^{H}-\hat{\bf I}^T\sqrt{N}{\bf R}{\bf F}^{H}
\end{eqnarray} 

By identifying the left hand to the right, we could obtain :
\begin{eqnarray}
\hat{\bf w} = \frac{\sqrt{N}{\bf F}^{H}(\rho\hat{\bf w}_{c}+\hat{\bf I}^T)}{N(\lambda_1+\rho)}
\end{eqnarray} 

Recalling the property of Fourier Transform that $\hat{x} = \sqrt{N}{\bf F}x$, we could derive the inverse Fourier  
Transform formula as: $\mathcal{F}^{-1}(\hat{x}) = \frac{1}{\sqrt{N}}{\bf F}^{H}\hat{x}$. In this case, we get the element-wise form of the filter in spatial domain eventually: 

\vspace{0.2cm} 
\begin{eqnarray}
 {\bf w}_r = {\bf r}\odot {\bf w} = {\bf r}\odot \frac{\mathcal{F}^{-1}(\rho\hat{\bf w}_{c}+\hat{\bf I}^T)}{\lambda_1+\rho} 
\end{eqnarray} 
%
%
%





%

%
\bibliographystyle{IEEEtran}   
\bibliography{IEEEabrv,mybibfile}
\end{document}